\documentclass[10pt,twocolumn,letterpaper]{article}

\usepackage[pagenumbers]{cvpr}              

\newcommand{\thickrule}{\specialrule{1.0pt}{0pt}{0pt}}

\usepackage{minted}

\definecolor{cvprblue}{rgb}{0.21,0.49,0.74}
\usepackage[pagebackref,breaklinks,colorlinks,allcolors=cvprblue]{hyperref}

\usepackage{makecell,array}
\usepackage[table]{xcolor}
\usepackage{colortbl}
\usepackage{amsmath}
\usepackage{url}
\usepackage{booktabs}
\usepackage{multirow}
\usepackage{amssymb,graphicx}
\usepackage{caption}
\usepackage{subcaption}
\usepackage{enumitem}
\usepackage{adjustbox}
\usepackage{xspace}
\usepackage{etoc}

\definecolor{darkorange}{rgb}{1.0, 0.55, 0.0}

\newcommand{\cellfirst}[1]{\textbf{#1}}
\newcommand{\cellsecond}[1]{\underline{#1}}
\newcommand{\cellthird}[1]{{#1}}

\newcommand{\apref}[1]{Appx.~\ref*{#1}}

\newcommand{\method}{EgoEdit\xspace}
\newcommand{\methoddmd}{EgoEdit-DMD\xspace}
\newcommand{\methodsf}{EgoEdit-RT\xspace}
\newcommand{\methodacronym}{EgoEdit\xspace}
\newcommand{\datasetname}{EgoEditData\xspace}
\newcommand{\benchmarkname}{EgoEditBench\xspace}

\newcolumntype{C}[1]{>{\centering\arraybackslash}p{#1}}

\newcommand{\generator}{\mathcal{G}}

\newcommand{\inputtensor}{\mathbf{X}}
\newcommand{\inputtensorsource}{\mathbf{X}^{src}}
\newcommand{\inputtensortarget}{\mathbf{X}^{tgt}}

\newcommand{\inputtensortargettime}{\mathbf{X}^{tgt}_{t}}
\newcommand{\inputtensortime}{\mathbf{X}_{t}}
\newcommand{\inputtensornoise}{\mathbf{X}_{0}}
\newcommand{\inputtensorclean}{\mathbf{X}_{1}}
\newcommand{\conditioning}{{c}}

\newcommand{\velocity}{{v}}
\newcommand{\velocitypred}{{\hat{v}}}

\newcommand{\datadistribution}{p_{d}}
\newcommand{\timestepdistribution}{p_{t}}
\newcommand{\noisedistribution}{p_{n}}
\newcommand{\difftimestep}{t}

\definecolor{mediumtealblue}{rgb}{0.0, 0.33, 0.71}
\definecolor{darkpastelgreen}{rgb}{0.01, 0.75, 0.24}
\definecolor{azure}{rgb}{0.0, 0.5, 1.0}

\definecolor{crimsonred}{rgb}{0.86, 0.08, 0.24}
\definecolor{firebrick}{rgb}{0.7, 0.13, 0.13}
\definecolor{carmine}{rgb}{0.59, 0.0, 0.09}
\definecolor{rosewood}{rgb}{0.4, 0.0, 0.04}
\definecolor{deepcherry}{rgb}{0.6, 0.13, 0.13}
\newcommand{\ours}{(ours)}

\def\paperID{xxx} 
\def\confName{xxx}
\def\confYear{2026}

\usepackage{xcolor}

\title{EgoEdit: Dataset,  Real-Time Streaming Model, and Benchmark \\ for Egocentric Video Editing}

\author{%
Runjia Li$^{1,3,\dagger}$\quad
Moayed Haji Ali$^{1,2}$\quad
Ashkan Mirzaei$^{1}$\quad
Chaoyang Wang$^{1}$\quad \\
Arpit Sahni$^{1}$\quad 
Ivan Skorokhodov$^{1}$
Aliaksandr Siarohin$^{1}$\quad
Tomas Jakab$^{3}$\quad
Junlin Han$^{3}$\quad \\
Sergey Tulyakov$^{1}$\quad
Philip Torr$^{3}$\quad
Willi Menapace$^{1}$\\[0.5em]
$^{1}$Snap Research
\qquad
$^{2}$Rice University 
\qquad
$^{3}$University of Oxford \\[0.1em]
\small \href{https://snap-research.github.io/EgoEdit/}{\texttt{snap-research.github.io/EgoEdit}}%
}

\begin{document}
\twocolumn[\maketitle\vspace{-2em}

    \centering
    \includegraphics[width=\textwidth]{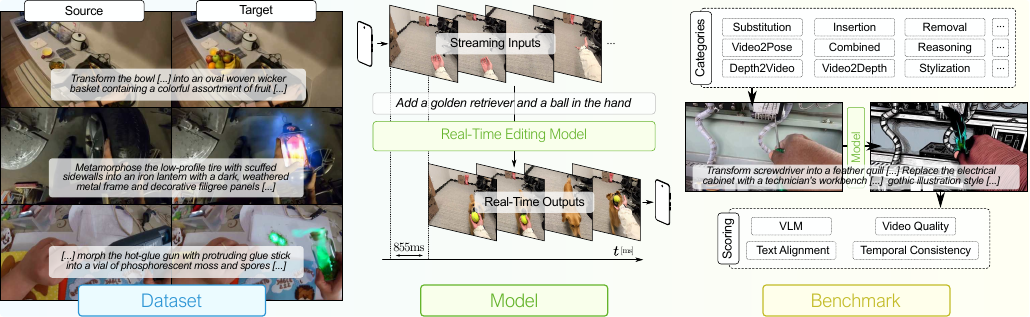}

    \captionof{figure}{We propose a framework for real-time egocentric video editing. Our system is composed of: \textbf{\datasetname{}}, a manually curated dataset of 100k video editing pairs focusing on the egocentric case and featuring object substitution and removal under challenging hand occlusions, interactions, and large egomotion; \textbf{\method{}}, the first real-time autoregressive model for egocentric video editing running in real time on a single H100 with 855ms first-frame latency and enabling live augmented reality (AR) interactions; \textbf{\benchmarkname{}}, a comprehensive benchmark for evaluation of egocentric video editing systems.
    }
    \label{fig:teaser}
\bigbreak]

\begingroup
\renewcommand\thefootnote{}\footnotetext{\textsuperscript{$\dagger$}Work done while interning at Snap Inc.}
\endgroup

\begin{abstract}
We study instruction-guided editing of egocentric videos for interactive AR applications. While recent AI video editors perform well on third-person footage, egocentric views present unique challenges — including rapid egomotion and frequent hand–object interactions — that create a significant domain gap. Moreover, existing offline editing pipelines suffer from high latency, limiting real-time interaction.
To address these issues, we present a complete ecosystem for egocentric video editing. First, we construct \datasetname{}, a carefully designed and manually curated dataset specifically designed for egocentric editing scenarios, featuring rich hand-object interactions, while explicitly preserving hands. Second, we develop \method{}, an instruction-following egocentric video editor that supports real-time streaming inference on a single GPU. Finally, we introduce \benchmarkname{}, an evaluation suite targeting instruction faithfulness, hand and interaction preservation, and temporal stability under egomotion.
Across both egocentric and general editing tasks, EgoEdit produces temporally stable, instruction-faithful results with interactive latency. It achieves clear gains on egocentric editing benchmarks—where existing methods struggle—while maintaining performance comparable to the strongest baselines on general editing tasks. \datasetname{} and \benchmarkname{} will be made public for the research community.

\end{abstract}    
\begin{figure*}[t]
    \centering
    \includegraphics[width=\linewidth]{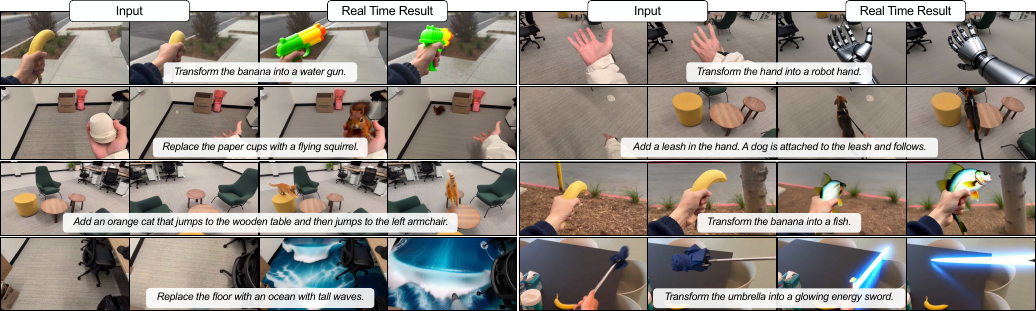}
    \caption{In-the-wild video edits produced in real time by \method's streaming variant \methodsf{} on a single H100 GPU. The model demonstrates strong generalization to out-of-distribution scenarios, producing compelling real-time results suitable for immersive AR experiences. Additional results are presented in \apref{appx:additional_results} and the website.}
    \label{fig:qualitatives_in_the_wild}
\end{figure*}
\begin{figure}
\includegraphics[width=\linewidth]{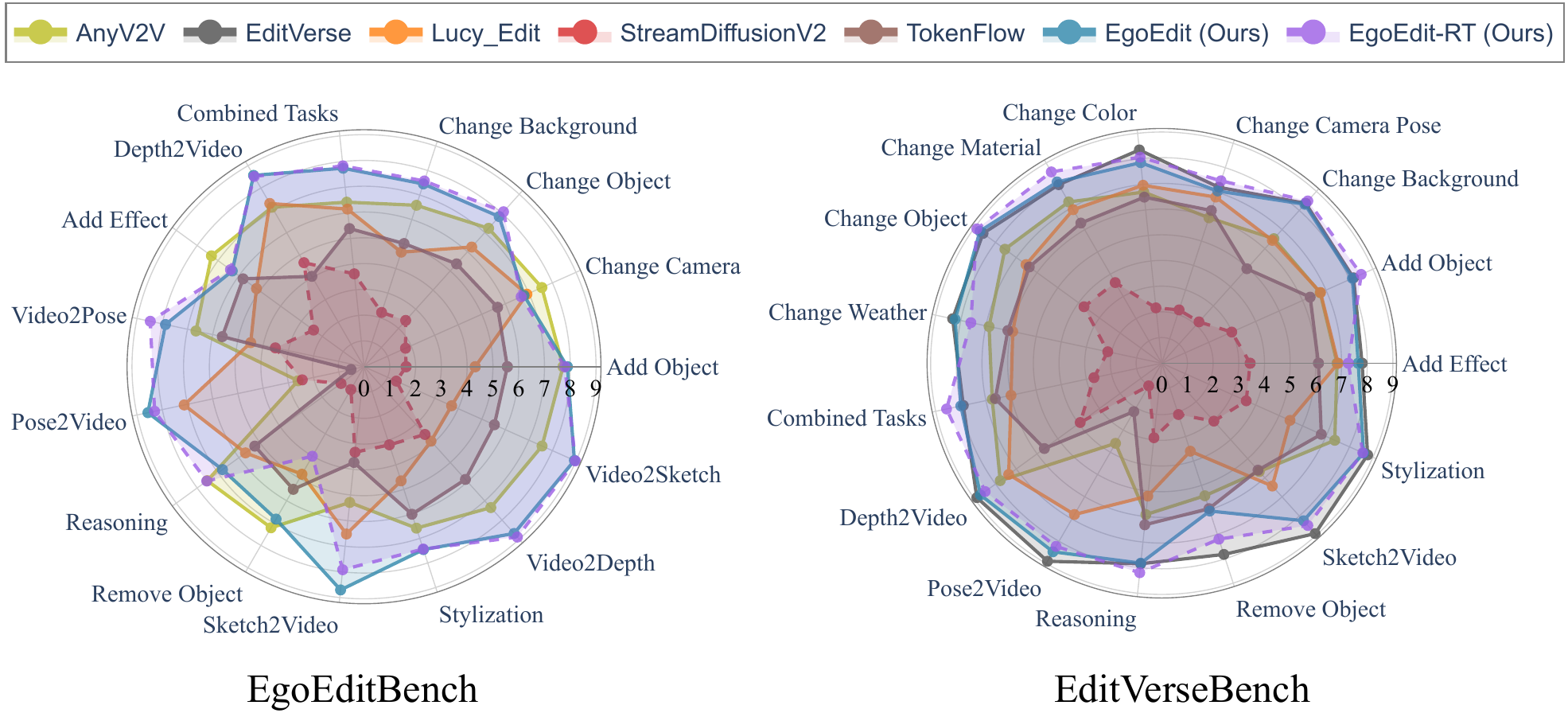}
\caption{Comparison of \method and \methodsf against baselines according to VLM score on \benchmarkname{} and EditVerseBench~\cite{ju2025editverse}. \method and its real-time variant \methodsf achieve superior results on egocentric editing tasks and perform competitively with the strongest baselines on general editing tasks. EditVerse is excluded from EgoEditBench as source code is unavailable. Streaming models are indicated in dashed lines.}
\label{fig:performace_per_task}
\end{figure}

\section{Introduction}

\label{sec:intro}

Altering the perceived world is central to augmented reality (AR). This technology empowers real-time experiences that immerse the user into new worlds, transform the surrounding environment, create virtual characters and objects, and let users \emph{interact} with these elements. However, traditional AR experiences rely on graphics pipelines and significant expert effort to handcraft each application, tying the potential achievable by this technology to the amount of available expert labor. The rapid progress of text-conditioned image~\cite{cao2025hunyuanimage30,labs2025flux1kontext,flux} and video~\cite{wan,kong2025hunyuanvideo,cogvideox} generation however raises a natural question: can instruction-guided \emph{editing}~\cite{brooks2022instructpix2pix,ju2025editverse,lucyedit2025,wang2025seededit30,wu2025qwenimage} serve as a direct engine for AR, enabling users to add, remove, or modify scene elements with simple language while they interact with the world?

Editing is especially appealing because it addresses central tasks at the core of AR, such as object insertion and removal, environmental restyling, camera and lighting adjustments, while preserving the core content of a scene. However, most recent editors~\cite{lucyedit2025,ju2025editverse} and their training corpora~\cite{zi2025senorita2m} are targeted at \emph{exocentric} content: third-person views with moderate motion, and low amounts of interaction. However, in AR, the camera is first-person and constantly moving, hands frequently occlude and manipulate objects, and object interactions are complex. These characteristics produce a distribution shift that limits the reliability of current editors~\cite{lucyedit2025} and the usefulness of existing datasets~\cite{zi2025senorita2m,wu2025insvie1m,yu2025veggie}. Moreover, AR requires not only edit fidelity but also \emph{real-time}, \emph{low-latency} responses suitable for \emph{interaction}. Many high-quality diffusion editing pipelines remain too slow for this setting.

We address this gap by targeting the egocentric editing regime end-to-end through data, a real-time generative model, and an evaluation framework as shown in~\Cref{fig:teaser}. \\ 
\noindent\textbf{Data.} High-quality paired editing data showing ``before/after'' editing examples are main drivers of editing quality~\cite{sheynin2023emuedit,liu2025step1xedit,ju2025editverse}. We therefore focus on \datasetname{}, a manually curated video editing dataset tailored to egocentric scenarios. Our data generation pipeline emphasizes the tasks most relevant to AR, focusing on the challenging removal and substitution of objects under active hand-object interaction, while explicitly preserving hand structure. We automatically identify actively interacted objects at which to target edits to ensure edit relevance, and conduct multi-stage human review to enforce consistently high visual quality. To ensure high instruction alignment, each pair is accompanied by precise, descriptive edit instructions. The resulting \datasetname{} dataset comprises 49.7k video samples and 99.7k instruction–edit pairs and will be released to support research in this domain.\\
\noindent\textbf{Real-Time Model.} Building on this data, we train an instruction-following egocentric editor, \method{}, from a large video generator. To meet the latency demands of interactive use, we apply Self-Forcing distillation~\cite{huang2025selfforcing} to obtain a generator that runs in real-time with low latency on a single H100 GPU, enabling responsive, user-in-the-loop editing as shown in \Cref{fig:qualitatives_in_the_wild}.\\
\noindent\textbf{Evaluation.} To standardize assessment in this setting, we introduce \benchmarkname{}, an automated benchmark for egocentric video editing. The benchmark targets instruction faithfulness, preservation of hands and manipulated objects, and temporal consistency under typical egocentric scenarios, providing a basis for reproducible comparisons.\\

\method{} delivers temporally stable, instruction-faithful edits, and latency suitable for interactive use. In comparisons against recent video-editing baselines~\cite{lucyedit2025,cheng2024insv2v,feng2025streamdiffusionv2} (see \Cref{fig:performace_per_task}), we observe consistent qualitative and quantitative improvements in the egocentric case, while performing closely to concurrent state-of-the-art editing methods~\cite{ju2025editverse} in the exocentric case. Ablations highlight the enabling role of \datasetname{} in achieving such performance.

In summary, we present a complete ecosystem for egocentric video editing comprised of:
\begin{itemize}
\item \textbf{\datasetname{}}: the first high-quality, manually curated dataset for egocentric video editing with 49.7k videos and 99.7k instruction–edit pairs intended for public release.
\item \textbf{\method{}}: a real-time egocentric video editing model that enables interactive AR scenarios on a single H100 GPU at 38.1fps with a first-frame latency of 855ms.
\item \textbf{\benchmarkname{}}: a benchmark that standardizes evaluation for egocentric editing with a focus on complex hand interactions and large egomotion typical of AR use cases.
\end{itemize}

\section{Related Work}
\label{sec:related}

\noindent\textbf{Image \& Video Editing.}
Text-conditioned diffusion initially enabled editing without additional training. Inversion-based methods reconstruct the source along the denoising trajectory and then steer it with the edit prompt~\cite{mokady2023nulltextinversion,lin2024scheduleedit,shen2024theblessingofrandomness,feng2025dit4edit,liu2024videop2p,qi2023fatezero,wang2024zeroshotvideoeditingusing,ju2024pnp}. Attention-control methods modify or reweight cross/self-attention to preserve content while changing appearance~\cite{hertz2022prompttoprompt,tumanyan2023plugandplay,cao2023masactrl,qi2023fatezero}. 
While broadly applicable, these approaches remain brittle for large structural edits and long-range consistency.
InstructPix2Pix~\cite{brooks2022instructpix2pix} overcame these limitations by collecting paired ``before/after'' image edit data, and training a conditional editor that directly edits the source image. 
Because paired video edit data are hard to obtain, early trained video editing approaches sidestepped supervision by coupling image editors with video modules~\cite{singer2025eve,liu2025genprop}. 
InsV2V~\cite{cheng2024insv2v}, however, brought the InstructPix2Pix recipe to video via synthetic video edit pairs.
Recent image editors improve fidelity by scaling both data and models, making use of large curated edit corpora~\cite{wu2025omnigen2,liu2025step1xedit} paired with either transformer backbones driven by MLLM embeddings~\cite{liu2025step1xedit,wang2025seededit30,cai2025hidreami1,wu2025qwenimage} or unified architectures that perform understanding and generation in one model for ``in-context'' editing~\cite{deng2025bagel,wu2025omnigen2,liao2025mogao}.
Video editors follow similar paradigms, with in-context editors unifying conditioning and instructions along the sequence: UNIC~\cite{ye2025unic} composes tasks via composite token sequences with task-aware RoPE and condition bias; EditVerse~\cite{ju2025editverse} improves unified image/video editing and generation through careful data curation. In contrast to image editors, sequence concatenation increases inference cost significantly for video editors~\cite{ju2025editverse}. Lucy Edit~\cite{lucyedit2025} reduces the cost of long sequences by channel-wise conditioning while retaining strong source-video control. 
Recent works aim at training-free real-time video editing~\cite{liang2024looking,kodaira2025streamdiffusion,feng2025streamdiffusionv2}. While showing promise, they currently suffer a quality gap with trained methods~\cite{lucyedit2025,ju2025editverse,ye2025unic}.

Across both images and videos, edit quality tracks the scale and quality of paired data and model capacity. This motivates the construction of our curated egocentric editing dataset, where edit examples focus on egocentric motion and complex hand/object interactions and are not available in existing video editing datasets.

\smallskip
\noindent\textbf{Editing Datasets.}
Editing quality closely follows the scale and curation of paired ``before/after'' data. For images, InstructPix2Pix~\cite{brooks2022instructpix2pix} established the data-first recipe with 313k CLIP-filtered pairs generated via Prompt-to-Prompt. Since then, most large corpora follow a common pattern: task-specialized synthetic pipelines that generate at scale~\cite{sheynin2023emuedit,ge2024seeddataedit,wei2025omniedit,yu2025anyedit,liu2025step1xedit,wang2025gptimageedit}, paired with multi-stage filtering using CLIP~\cite{yu2025anyedit,ge2024seeddataedit}, MLLMs~\cite{wei2025omniedit,yu2025anyedit}, object detectors~\cite{yu2025anyedit}, heuristics, and human-in-the-loop passes~\cite{ge2024seeddataedit,wei2025omniedit,yu2025anyedit,liu2025step1xedit,wang2025gptimageedit}, resulting in million-scale filtered datasets. Quality-focused variants tighten these stages with heavier MLLM and human curation~\cite{liu2025step1xedit} or use existing high-quality editing methods as data generators~\cite{wang2025gptimageedit}. Complementary, carefully curated sets show that domain alignment and quality can result in improvements despite smaller sizes~\cite{zhang2023magicbrush,zeng2025editworld}.
For videos, the same principles now dominate. InsV2V~\cite{cheng2024insv2v} pioneered synthetic video edit pairs, demonstrating that scaled paired training transfers to the temporal setting. Follow-ups are based on propagation of edited frames~\cite{wu2025insvie1m,yu2025veggie} or the creation of specialized pipelines mirroring the image editing domain~\cite{zi2025senorita2m}. EditVerse~\cite{ju2025editverse} extensively filtered existing datasets~\cite{zi2025senorita2m,wei2025omniedit,yu2025anyedit} resulting in a curated set of 232K video-edit samples.  

\smallskip
\noindent\textbf{Streaming Video Generation.}
State‑of‑the‑art video generators~\cite{wan,cogvideox,kong2025hunyuanvideo} yield high-quality but suffer from low throughput, long first-frame latency, and limited clip length. The denoising process is long, and the full video  must be generated before the first frame can be shown to the user. Recent methods create autoregressive generators capable of generating long videos by predicting a chunk of frames at a time. Diffusion forcing~\cite{chen2024diffusionforcing} and its variants~\cite{chen2025skyreelsv2,ai2025magi1,li2025vmem, xiao2025worldmemlongtermconsistentworld} divide the video into chunks and assigning distinct diffusion noise levels so the model can denoise autoregressively chunk‑by‑chunk. 
Causal distillation converts slow bidirectional teachers into few‑step causal students. CausVid~\cite{yin2025causvid} distills a 50‑step bidirectional model into a 4‑step causal student using DMD~\cite{yin2024dmd2}, with chunk‑wise generation and KV caching. Self‑Forcing~\cite{huang2025selfforcing} addresses the exposure bias by rolling out the student at train time, allowing the model to self-correct its errors. APT2~\cite{lin2025apt2} employs adversarial post‑training to achieve 1‑NFE per frame for interactive generation. Our approach follows the Self‑Forcing strategy to achieve interactive egocentric video editing. 

\section{\datasetname}
\label{sec:dataset}

Compared to conventional video editing, egocentric editing brings distinct challenges due to complex hand-object interactions, frequent occlusions, and large egomotion. Existing editing datasets rarely cover egocentric scenes or such intricate interactions, creating data barriers~\cite{zeng2025editworld} for learning-based AR experiences.
To address data scarcity and provide a foundation for egocentric editing, we present \datasetname{}, a manually curated and high-quality egocentric video editing dataset, focusing on rich hand-object interactions.

\subsection{Data Curation Pipeline}

Because data quality~\cite{liu2025step1xedit,sheynin2023emuedit,zhang2023magicbrush} and domain alignment~\cite{zeng2025editworld} are main drivers of editing performance, our pipeline prioritizes quality over quantity through strict filtering, and emphasizes creation of editing pairs that depict the most challenging egocentric scenarios with rich hand-object interaction. Starting from real egocentric videos~\cite{grauman2022ego4d,grauman2024egoexo4d}, we ensure that:
(1) videos contain an object that is actively manipulated by the egocentric subject,
(2) the edit target is the manipulated object,
(3) the synthetic videos preserve realistic hand motion while reflecting the intended content change,
(4) the instruction prompts are descriptive and accurate. 
The stages are detailed below, with additional implementation specifics in \apref{appx:pipeline_details}.

\noindent\textbf{Video selection.}
We consider videos from the Ego4D~\cite{grauman2022ego4d} and EgoExo4D~\cite{grauman2024egoexo4d} datasets. From Ego4D we select sequences coming from high-quality camera models only (see \apref{appx:pipeline_details}). For EgoExo4D, we select egocentric cameras, and perform rectification of the videos. We also conduct filtering to reduce jittering and motion blur. 1.8\% of videos are retained after this stage.

\noindent\textbf{Hand mask segmentation.}
We first detect hands in each frame using a hand detection method~\cite{potamias2024wilor}. Videos without visible hands are removed. For the remaining videos, detected hand regions provide visual prompts to SAM 2~\cite{ravi2025sam}, which yields fine-grained and temporally consistent hand masks across the sequence. We conduct human filtering to ensure hand masks are annotated correctly, with 49.6\% of samples remained after filtering.

\noindent\textbf{Object names extraction.}
We then identify the object that the hands manipulate using a vision language model.
Qwen2.5-VL-32B~\cite{bai2025qwen25vl} is prompted to name the interacted objects for each video. Videos where no meaningful hand-object interaction is found are discarded.

\noindent\textbf{Object mask segmentation.}
Given the identified object name, Grounded SAM~\cite{ren2024groundedsam} predicts an approximate object mask in each frame. Videos with low mask confidence in every frame are removed. To confirm real interaction, we compute the edge distance between the hand and object masks and the distances between the object mask and hand skeleton keypoints~\cite{potamias2024wilor}, and filter out false positives. The coarse grounded masks then seed SAM 2~\cite{ravi2025sam} to obtain fine-grained and temporally consistent object masks. We conduct manual filtering to ensure object masks are extracted correctly, retaining  43.6\% of the sequences. Curation at this stage ensures the expensive successive object editing stage is only run on correctly processed videos.

\noindent\textbf{Object Editing.}
This stage creates target videos where the original object is replaced with a different one or removed entirely.
For each segment, we prompt GPT-5 Mini~\cite{gpt5} to propose diverse target objects for substitution, including both ordinary and imaginary items. Qwen-Image~\cite{wu2025qwenimage} then synthesizes a reference image for each proposal. Next, GPT-5 Mini~\cite{gpt5} produces a scene-level description of the video assuming interaction with the target object. We feed the reference image, the scene-level prompt, and the object mask to Wan 2.1 VACE 14B~\cite{jiang2025vace} to generate the edited video. Object removal is treated as a special case with no target object.
Although the conditioning is rich and the computation is heavy ($0.112$ fps on 8 H100 GPUs), Wan 2.1 VACE 14B~\cite{jiang2025vace} yields only a small fraction of results that meet our standard for dataset quality. Human annotators therefore remove imperfect results to ensure consistent quality, leaving 37.8\% of the generated edits. 

\noindent\textbf{Editing pairs construction.}
We form video editing pairs that include a source video, a target video, and a natural language instruction. For each real video and its edited variants, we permute pairs among all versions, including the original clip, and prompt GPT-5 Mini~\cite{gpt5} to generate a precise and faithful description of the edit~\cite{wang2025gptimageedit}.

\subsection{Statistics}
After stages of curation and filtering, only 0.4\% of the original videos from Ego4D and EgoExo4D are kept. The resulting pairs compose our \datasetname{} dataset. It is composed of 10.9k original and 38.8k synthetic videos (70 hours long), for an average of 3.6 synthetic videos per original video, yielding a total of 99.7k editing pairs, each comprising a source video, target video, and editing instruction, with 93,422 of them derived from Ego4D~\cite{grauman2022ego4d} and 6,237 pairs from EgoExo4D~\cite{grauman2024egoexo4d}. 
\datasetname{} is diverse.
Additional details are presented in \apref{appx:dataset_details}.

\section{EgoEdit}
\label{sec:method}

\begin{figure}[t]
    \centering
    \includegraphics[width=\linewidth]{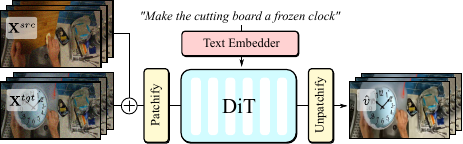}
    \caption{\textbf{Architecture of \methodacronym{}.} \methodacronym{} extends a video generation DiT model for video editing by performing channel-wise concatenation of the source and noisy target video inputs, avoiding the computational overheads of sequence-wise concatenation.}
    \label{fig:architecture_training}
\end{figure}

Consider a source video $\inputtensorsource$ and a textual instruction $\conditioning$ specifying a desired edit. The goal is to produce a target video $\inputtensortarget$ reflecting the requested change.
We focus on egocentric video editing. Unlike the traditional exocentric setting, which can be processed offline, 

egocentric editing requires real-time results to support interactive experiences. 

To address these challenges, we introduce \method, a video editing method tailored to the real-time egocentric setting.
We first convert a pretrained video generator into a video editor by adding source video conditioning and finetuning on an editing corpus that includes \datasetname.

Then, we distill the editor into an autoregressive real-time generator using bidirectional DMD~\cite{yin2024dmd2} and autoregressive Self Forcing~\cite{huang2025selfforcing}.
\Cref{sec:flow_matching} introduces the Flow Matching framework,
\Cref{sec:architecture} describes the model, and
\Cref{sec:distillation} details the distillation procedure.

\subsection{Preliminaries: Flow Matching}
\label{sec:flow_matching}

We train our generators with Rectified Flow flow matching~\citep{liu2022rectifiedflow,lipman2023flowmatching}, which learns a deterministic path from a noise distribution $\noisedistribution$ to the data distribution $\datadistribution$.
Let $\inputtensorclean \sim \datadistribution$ and $\inputtensornoise \sim \noisedistribution=\mathcal{N}(0,\mathbf{I})$.
We define a linear path
$\inputtensor_{\difftimestep} \;=\; (1-\difftimestep)\,\inputtensornoise + \difftimestep\,\inputtensorclean$ for $\difftimestep \in [0,1]$,
whose ground truth velocity is constant along the path,
$\velocity_{\difftimestep} = \frac{d\inputtensor_{\difftimestep}}{d\difftimestep} = \inputtensorclean-\inputtensornoise$.
A neural network $\generator(\cdot)$ predicts the velocity from a noised input and a time value and is trained by minimizing:
\begin{equation}
\label{eq:rf}
\resizebox{0.9\linewidth}{!}{$
\mathcal{L}_{\text{RF}}
= \mathbb{E}_{\difftimestep \sim \timestepdistribution,\,\inputtensorclean \sim \datadistribution,\,\inputtensornoise \sim \noisedistribution}
\big\|\,\generator(\inputtensor_{\difftimestep},\,\difftimestep) \!-\! (\inputtensorclean\!-\!\inputtensornoise)\,\big\|_2^2,
$}
\end{equation}
where $\timestepdistribution$ is a training distribution over $\difftimestep$ chosen as a logit-normal~\citep{sd3}.
At inference time, an Euler solver integrates the learned ODE from $\inputtensornoise$ to $\inputtensorclean$ to produce a sample.

\subsection{Architecture}
\label{sec:architecture}

We base \method{} on a pretrained text-to-video generator trained on the latent space of a Wan 2.1 autoencoder~\cite{wan} with a transformer backbone~\cite{dit} (see \Cref{appx:text_to_video_model}).
As shown in \Cref{fig:architecture_training}, the model receives $\inputtensortime$ at time $\difftimestep$, projects it to a sequence of tokens through a linear patchifier, processes the tokens with transformer blocks, and maps them to the predicted velocity $\velocitypred$ through a final linear head.
Text conditions $\conditioning$ are provided through cross attention layers placed after each self attention block. The computation is expressed as $\velocitypred=\generator(\inputtensortime\mid\conditioning)$.

We adapt the model to editing by replacing $\inputtensortime$ with the noisy target video $\inputtensortargettime$ and by conditioning on the source video $\inputtensorsource$, written as $\velocitypred=\generator(\inputtensortargettime\mid\inputtensorsource;\conditioning)$.
We consider two main strategies to inject $\inputtensorsource$.
Sequencewise concatenation~\cite{ju2025editverse,ye2025unic} patchifies the source and concatenates its patches with those of the target along the sequence dimension.
This approach is common~\cite{ju2025editverse,wu2025qwenimage}, but the longer token sequence increases the cost of self attention quadratically, which conflicts with real-time low latency operation.
To avoid this growth, \method{} uses channel-wise concatenation~\cite{lucyedit2025}, where $\inputtensorsource$ and $\inputtensortargettime$ are concatenated along channels before patchification, which keeps the cost close to the base model.

\begin{figure}[t]
    \centering
    \includegraphics[width=0.9\linewidth]{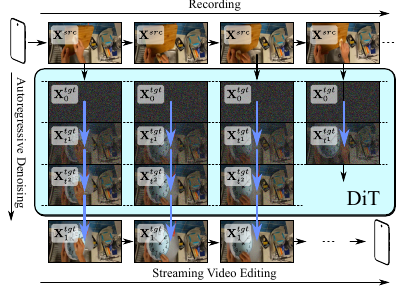}
    \caption{\textbf{Inference of \methodacronym{}.} \methodacronym{} performs inference in a streaming fashion. A camera continuously acquires video sequences which are edited by the model in a chunk-by-chunk manner so that the edited video can be served to the user in a watch-as-you-generate fashion. Each blue arrow represents a model forward pass on a single video chunk for the case of a 3 steps model.}
    \label{fig:architecture_inference}
\end{figure}
\begin{figure*}
    \centering
    \includegraphics[width=\linewidth]{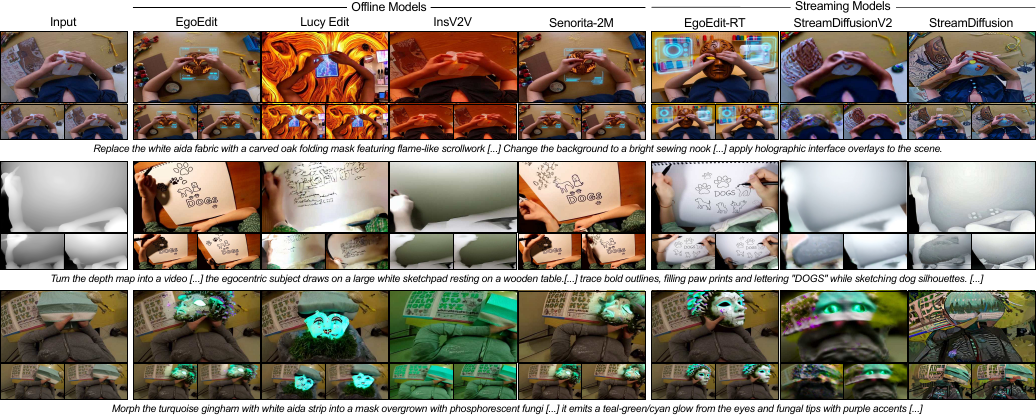}
    \caption{Qualitative comparison of \method{} and its real time streaming variant \methodsf{} against baselines on \benchmarkname{}. \method and \methodsf consistently perform better than their baselines. Note that Señorita-2M uses the first frame from \method for frame propagation. Additional results are presented in \apref{appx:additional_results} and the website.}
    \label{fig:qualitatives_baselines}
\end{figure*}

\subsection{Distillation Procedure}
\label{sec:distillation}

The edited model described above is accurate but slow at inference.
40 denoising steps with classifier-free guidance are required to produce a video, which corresponds to 80 model invocations (NFEs). In addition, the model generates the full clip at once, which delays the first visible frame until the process finishes.
Interactive use calls for autoregressive generation with low latency, as illustrated in \Cref{fig:architecture_inference}.
We therefore distill the editor into a real-time autoregressive model in two phases (see \apref{appx:distillation_procedure}). 

\noindent\textbf{Bidirectional DMD distillation.}
We follow DMD~\cite{yin2024dmd2} to compress the 40-step model with classifier-free guidance into a 4-step model with distilled guidance.
This reduces the NFEs from 80 to 4, facilitating subsequent distillation.

\noindent\textbf{Self Forcing.}
Self Forcing~\cite{huang2025selfforcing} runs the causal model autoregressively on video streams and applies a DMD loss with score models based on the bidirectional teacher.
The model learns to correct its own errors, which reduces exposure bias and enables low latency autoregressive inference. To minimize exposure bias while achieving low latency, we generate a chunk at a time, where each chunk is composed of three latent frames. Note that the employed Wan~\cite{wan} autoencoder natively supports autoregressive operation.

\Cref{tab:latency_throughput_performance} shows latency and throughput for the different model variants. Our Self Forcing distilled model can produce results in real-time at 38.1fps with a latency of 855ms for displaying the first frame on a single H100 GPU.

\subsection{EgoEditBench}

\noindent Existing video editing benchmarks~\cite{ju2025editverse,chen2025ivebench,sun2024bench} are primarily built on third-person natural videos, making them unsuitable for evaluating egocentric video editing. To properly assess this setting, we follow~\cite{ju2025editverse} to construct \benchmarkname{}, a benchmark designed to evaluate editing performance across 15 egocentric tasks~\cite{ju2025editverse}, including: Add Object, Add Effect, Remove Object, Change Object, Change Background, Change Camera Pose, Stylization, Reasoning, Depth-to-Video, Sketch-to-Video, Pose-to-Video, Video-to-Pose, Video-to-Sketch, Video-to-Depth, and a Combined Task of all previous tasks. To build this benchmark, we sample 100 unique source videos from a split of the Ego4D dataset~\cite{grauman2022ego4d} that was not used in constructing \datasetname{}, ensuring maximum diversity. This is achieved by first extracting source-object names following the \datasetname{} pipeline, and computing BERT~\cite{devlin2019bert} embeddings from the concatenation of each source-object name and its corresponding scene description. We then perform K-means clustering with 10 centroids, and select 10 samples per cluster, resulting in 100 diverse source videos. Conditioned on the source video, its caption, and the source object, we prompt GPT-5~\cite{gpt5} to generate task-specific instruction prompts for each of the 15 tasks. For the X-to-Video tasks, we synthesize conditioning signals as follows: Canny edge maps using OpenCV for the Sketch2Video task, 2D poses using DWpose~\cite{yang2023dwpose} for Pose2Video, and depth maps using Depth Anything~\cite{depthanything} for Depth2Video. For Add Object and Remove Object, we sample 50 unique source videos per task and construct the corresponding instruction prompts following \datasetname{}. For the Change Object task, we generate four instruction prompts for each unique source video with two focusing on object replacement and two combining object replacement with an added effect.

In total, \benchmarkname{} comprises $1700$ source videos paired with instruction prompts, covering 15 diverse egocentric editing tasks. Evaluation is performed according to the EditVerseBench~\cite{ju2025editverse} protocol and metrics, with results averaged per task to ensure equal weighting across all tasks. 

Additional implementation details for \benchmarkname{} can be found in \apref{appx:benchmark_details}.

\section{Experiments}
\newcommand{\oursrowcolor}{azure!6}
\newcommand{\pickscorefull}{Pick Score\xspace}
\newcommand{\pickscoreshort}{PS\xspace}
\newcommand{\vlmevalfull}{VLM evaluation\xspace}
\newcommand{\vlmevalshort}{VLM\xspace}
\newcommand{\secrow}[1]{ & \multicolumn{8}{c}{\cellcolor{gray!12}\textbf{#1}}\\} 
\begin{table*}[t]
\centering
\small
\begin{tabular}{lccccccccc}
\toprule
\multirow{2}{*}{\vspace{-0.5em}\textbf{Method}}
& \multirow{2}{*}{\vspace{-0.5em}\textbf{Family}}
& \multicolumn{4}{c}{\textbf{EgoEditBench}}
& \multicolumn{4}{c}{\textbf{EditVerseBench}~\cite{ju2025editverse}} \\
\cmidrule(lr){3-6}\cmidrule(lr){7-10}
& & \textbf{\vlmevalshort} $\uparrow$ & \textbf{\pickscoreshort} $\uparrow$ & \textbf{TA} $\uparrow$ & \textbf{TC} $\uparrow$
& \textbf{\vlmevalshort} $\uparrow$ & \textbf{\pickscoreshort} $\uparrow$ & \textbf{TA} $\uparrow$ & \textbf{TC} $\uparrow$ \\
\midrule

TokenFlow~\cite{qu2025tokenflow} & \multirow{2}{*}{Attention manipulation} & 4.99 & \cellsecond{18.91} & \cellthird{15.89} & \cellthird{95.04} & 5.87 & \cellfirst{19.90} & 23.68 & 98.21 \\
STDF~\cite{yatim2024stdf}            & & 4.59 & 18.69 & 15.64 & 93.96 & 6.64 & 19.54 & \cellthird{24.33} & 96.96 \\

\midrule
$\text{Señorita-2M}^{^{\ddagger}}$~\cite{zi2025senorita2m}
& \multirow{2}{*}{First-frame propagation} & \cellsecond{7.52} & \cellthird{18.85} & \cellsecond{16.25} & \cellsecond{95.86} & \cellthird{6.99} & 19.32 & 23.07 & 98.33 \\
$\text{AnyV2V}^{^{\ddagger}}$~\cite{ku2024anyv2v}          & & \cellthird{6.72} & 18.65 & 15.35 & 92.37 & 6.46 & 19.47 & 23.32 & 95.91 \\

\midrule
InsV2V~\cite{cheng2024insv2v}          & \multirow{4}{*}{Instruction-guided} & 5.24 & 18.81 & 14.92 & 94.01 & 5.71 & 19.08 & 22.49 & 96.39 \\
Lucy Edit~\cite{lucyedit2025}       & & 5.44 & 18.87 & 15.03 & 94.41 & 6.27 & 19.23 & 22.55 & \cellsecond{98.62} \\
$\text{EditVerse}^{\dagger}$~\cite{ju2025editverse}       & & --- & --- & --- & --- & \cellfirst{8.26} & \cellsecond{19.69} & \cellfirst{25.29} & \cellfirst{98.68} \\
\rowcolor{\oursrowcolor}
\textbf{\method~\ours}
& & \cellfirst{7.76} & \cellfirst{19.21} & \cellfirst{16.89} & \cellfirst{96.70}
& \cellsecond{8.00} & \cellthird{19.61} & \cellsecond{24.40} & \cellthird{98.54} \\

\thickrule

StreamDiffusion~\cite{kodaira2025streamdiffusion}   & \multirow{3}{*}{Streaming models} & \cellsecond{4.32} & \cellsecond{18.92} & \cellsecond{14.15} & 86.83 & \cellsecond{4.33} & \cellsecond{18.76} & \cellfirst{19.01} & 93.41 \\
StreamDiffusionV2~\cite{feng2025streamdiffusionv2} & & 2.55 & 18.63 & 12.75 & \cellsecond{94.31} & 2.78 & 18.45 & 17.32 & \cellsecond{98.22} \\
\rowcolor{\oursrowcolor}
\textbf{\methodsf~\ours}
& & \cellfirst{7.71} & \cellfirst{19.13} & \cellfirst{16.34} & \cellfirst{96.41} & \cellfirst{8.18} & 
  \cellfirst{19.59} & \cellsecond{17.61}  &  \cellfirst{98.55}\\
\bottomrule
\end{tabular}
\caption{Quantitative comparison of the baseline models and our method on EgoEditBench and EditVerseBench benchmarks: ``VLM'' is VLM evaluation score, ``PS'' is Pick Score, ``TA'' is Text Alignment, ``TC'' is Temporal Consistency. Reference-based editing tasks from EditVerseBench—propagation, inpainting, reference insertion, and edit with mask---were excluded. ``$\dagger$'' indicates closed-source models evaluated using their publicly released samples; ``$\ddagger$'' indicates models utilizing the first frame generated by \method. \methodsf stands for the real-time streaming version of \method.}
\label{tab:egobench}
\end{table*}

\label{sec:experiments}



\subsection{Experimental Setup}

\noindent\textbf{Training details.} We finetune our pretrained video generation model on \datasetname{} and a corpus of additional 1.31M video and 3.5M image editing pairs to obtain the base video editing model. We finetune the model with a total batch size of 96 for 30k iterations using an AdamW~\cite{adamw} optimizer with lr 1e-5, weight decay of 0.1 and exponential moving average. Bidirectional DMD distillation is performed for 4.5k steps using an AdamW~\cite{adamw} optimizer with lr of 1e-6 for the model and 4e-7 for the critic, weight decay of 0.1 and exponential moving average. Successively, Self Forcing training is conducted for 3.5k steps using an AdamW~\cite{adamw} optimizer with lr of 1e-6 for the model and 4e-7 for the critic, weight decay of 0.1 and exponential moving average, producing the final checkpoint.
Additional training and dataset details are presented in the \apref{appx:training_details}.


\noindent\textbf{Evaluation metrics and protocol.} \benchmarkname{} serves as the main framework for evaluating egocentric video editing quality. Evaluation is supplemented by EditVerseBench~\cite{ju2025editverse} to provide references on the non-egocentric case. As our model does not support reference image conditioning, we remove the EditVerseBench tasks of Propagation, Inpainting, Reference Insertion, and Edit with Mask, which require such conditioning.

\subsection{Comparison to Baselines}

\noindent\textbf{Baselines selection.} We select a range of baselines against which to compare that are based on attention manipulation~\cite{qu2025tokenflow,yatim2024stdf}, propagation of the first frame~\cite{zi2025senorita2m,ku2024anyv2v}, or direct instruction-guided video to video translation~\cite{ju2025editverse,lucyedit2025,cheng2024insv2v}.
Frame propagation methods~\cite{zi2025senorita2m,ku2024anyv2v} receive as input the first frame edited by \method for fair comparison.
Additionally, we select StreamDiffusion~\cite{feng2025streamdiffusionv2,kodaira2025streamdiffusion} as representatives for real-time editing methods. For a fair comparison, all baseline models are evaluated using their recommended inference settings, including the number of sampling steps, guidance scale, resolution, and frame rate (following EditVerseBench~\cite{ju2025editverse}). For inversion-based methods that require a target video caption and cannot process natural instruction prompts, we use GPT-5~\cite{gpt5} to generate the target caption from the source prompt and instruction prompt. Complete inference details for each baseline are provided in \apref{appx:baseline_details}.

\noindent\textbf{Results.} As shown in \Cref{tab:egobench} and \Cref{fig:qualitatives_baselines}, \method produces state-of-the-art results on egocentric videos while still performing strongly on general editing, as demonstrated by EditVerseBench performance. In particular, on the challenging egocentric setting, our method drops only 0.24 points in VLM evaluation when switching from general editing tasks to egocentric ones, while Lucy Edit~\cite{lucyedit2025} and InsV2V drop respectively 0.83 and 0.47 points. Only Señorita-2M~\cite{zi2025senorita2m} and AnyV2V~\cite{ku2024anyv2v} retain their performance due to receiving the propagated first frame from \method.
When compared to existing real-time streaming editors, our streaming variant \methodsf achieves markedly stronger performance across both benchmarks. Relative to the bidirectional full model, \methodsf delivers comparable results on all quantitative metrics. 
These findings highlight the strong real-time editing capabilities of \methodsf and validate the effectiveness of our distillation procedure. Additional qualitative results are provided in the supplementary material and the website.



\subsection{Ablations}
We conduct ablations to investigate performance change during distillation and the effectiveness of \datasetname{}.

\noindent\textbf{Distillation.}
\Cref{tab:latency_throughput_performance} compares the original non-distilled \method model to the one obtained after 4-step DMD distillation and the final real-time autoregressive generator obtained after Self Forcing training. We take \emph{first chunk latency} as the main metric for assessing real-time method suitability. It shows the delay from the moment the user hits the ``record'' button on their camera to the moment the corresponding first edited frame is visualized on the screen. Due to the inability of standard methods to perform chunk-by-chunk generation, only the Self Forcing variant achieves sub-second latency compatible with interactive usage. When evaluated on \benchmarkname{}, the Self Forcing variant reaches comparable VLM scores to the bidirectional teacher model, while enabling interactive generation. 

\noindent\textbf{Contribution of \datasetname{}.}
To better quantify how \datasetname{} improves a model’s ability to adapt to egocentric editing, we vary the number of videos of \datasetname{} included during training. Starting from the same text-to-video checkpoint, we finetune different models for 10k iterations on our training editing data corpora, removing a certain percentage of original videos and corresponding edited versions from \datasetname{} in a proportion from 0\% to 100\%. 
As shown in \Cref{tab:ablation}, performance on \benchmarkname{} steadily improves as more samples from \datasetname{} are incorporated, highlighting the role of \datasetname{} in enabling robust egocentric video editing. We also discuss how \datasetname{} influences general editing performance on EditVerse in \apref{appx:evaluation_details}.

\begin{table}  
\centering
\setlength{\tabcolsep}{3.0pt}
\small

\begin{tabular}{lccc}
\toprule

& \textbf{No Distill.} & \textbf{DMD}~\cite{yin2024dmd2} & \textbf{Self Forcing}~\cite{huang2025selfforcing} \\
\midrule
Is streaming? & $\times$ & $\times$ & \checkmark \\
NFEs & 80 & 4 & 4 \\
VLM-Eval~$\uparrow$ & 7.76 & 7.31 & 7.71 \\
First chunk size & 81 frames & 81 frames & 9 frames \\
Next chunk size & N$/$A & N$/$A & 12 frames \\
\midrule
& \multicolumn{3}{c}{\emph{First Chunk Latency [ms]}} \\
\midrule
Recording & 5062 & 5062 & 562 \\
AE & 1520 & 1520 & 217 \\
Model & 6850 & 343 & 75.7 \\
Total & 13432 & 6925 & 855 \\
\midrule
 & \multicolumn{3}{c}{\emph{Throughput [fps]}} \\
 \midrule

Model & 11.9 & 237 & 134 \\
Model + AE & 9.68 & 43.5 & 38.1 \\
\bottomrule

\end{tabular}
\caption{\benchmarkname{} VLM score, latency and throughput analysis of different  distilled \method models on 1$\times$H100 under resolution of $512\!\times\!384$px. We consider latency involved in recording the source video, running \method, and running the autoencoder (AE) for source video encoding and generated video decoding.}
\label{tab:latency_throughput_performance}
\end{table}



\begin{table}[h]
\centering
\small
\begin{tabular}{l|cccc}
\toprule
\rowcolor{gray!12}
\textbf{\% of \datasetname{}} & 0\% & 25\% & 75\% & 100\% \\
\midrule
VLM Evaluation$\uparrow$ & 4.87  &  7.12 & 7.52 & 7.85 \\
\bottomrule
\end{tabular}
\caption{Performance of \method{} when trained using progressively smaller subsets of \datasetname{}. A trend is visible: the model performs better with more egocentric editing data included during training. Note that these results differ from \Cref{tab:egobench} because all models are evaluated at the 10k-iteration checkpoint. Additional details are provided in \apref{appx:evaluation_details}.}
\label{tab:ablation}
\end{table}

\subsection{In-the-Wild Evaluation}
We conduct in-the-wild evaluation of the real-time version of \method{} to test robustness in the real-world usage and emerging capabilities. Results are presented in \Cref{fig:qualitatives_in_the_wild}. We observe that the model is able to perform complex editing tasks such as correctly preserving hands during interactions, modeling environment interactions such as the sidewalk becoming wet when sprayed with an imaginary water gun, modeling lighting effects induced by inserted objects, replacing fast-moving thrown objects, inserting animals that realistically interact with the environment by jumping over or navigating around objects, and correctly rotating inserted objects according to the substituted object orientation. Most excitingly, some instances of inserted objects react to user interactions, such as dogs being walked on a leash. We observe, however, that the amount of structural modifications that inserted objects can operate on the surrounding environment is limited: swords will not cut through furniture, and animals will not move real objects.

\section{Discussion}
\label{sec:discussion}

\noindent\textbf{Limitations.}
While \methodsf{} performs comparably to \method{} on automated benchmarks and presents strong in the wild generation capabilities, we notice a qualitative gap which manifests as: (i) lower proficiency in out-of-distribution editing instructions, (ii) less robust performance when editing objects becoming temporarily occluded, (iii) lower temporal consistency.
\method{} possesses a first-frame latency of 855ms, which is sufficient but suboptimal for interactive usage. As shown in \Cref{tab:latency_throughput_performance}, such latency is dominated by the recording time of the first chunk of 3 latent frames, corresponding to 9 RGB frames. Further reduction in latency can be achieved by lowering the chunk size during Self Forcing training.
Finally, \method{} operates at a resolution of $512\!\times\!384$px and a frame rate of 16 fps, slightly lower than the common 480p resolution.





\noindent\textbf{Conclusions.} We introduce a comprehensive framework for developing learning-based AR applications through egocentric video editing. To support this effort, we construct and publicly release \datasetname{}, the first high-quality and manually curated dataset of egocentric video edits, containing 99.7k editing pairs across 49.7k unique videos. Building on this dataset, we propose \method{}, the first real-time video editing model tailored to egocentric applications. Our model demonstrates strong generalization to in-the-wild videos, achieves state-of-the-art performance on egocentric editing tasks, and competitive results on general video editing benchmarks. Finally, we introduce \benchmarkname{}, a standardized benchmark designed to evaluate egocentric video editing performance under realistic hand–object interactions. Together, these contributions form a foundation and ecosystem for real-time, instruction-guided AR generation, paving the way for future research in interactive generative systems for augmented reality.
{
    \small
    \bibliographystyle{ieeenat_fullname}
    \bibliography{main}
}

\clearpage
\appendix

\part*{Appendix}
\noindent For more video results, please check the website.
\addcontentsline{toc}{part}{Appendix}   

\begingroup
  \etocsetnexttocdepth{subsection}     
  \etocsettocstyle{\section*{Appendix Contents}}{}
  \localtableofcontents
\endgroup


\section{Additional Dataset Details}
\label{appx:dataset_details}

\subsection{Pipeline Details}
\label{appx:pipeline_details}


\noindent\textbf{Video selection.} We select videos from Ego4D~\cite{grauman2022ego4d} and EgoExo4D~\cite{grauman2024egoexo4d} according to the following criteria. The video must be captured with one of the following camera models: GoPro Hero 4, GoPro Hero Black 7, GoPro Hero Black 8, GoPro Hero Black 9, GoPro Hero Silver 7, or GoPro Max, and it must be monocular rather than binocular. Furthermore, to ensure that the videos are sharp and visually informative, we apply additional filtering based on jitter scores and aesthetic scores.


\noindent\textbf{Hand mask segmentation.} We employ WiLoR~\cite{potamias2024wilor} for hand detection, applying a confidence threshold of 0.75 on a frame-by-frame basis. From the detected frames, we select the three with the highest confidence scores and use their hand masks to generate point prompts for SAM2~\cite{ravi2025sam}, enabling the creation of dense and temporally consistent hand masks across frames.

\noindent\textbf{Object names extraction.} We employ Qwen2.5-VL-32B~\cite{bai2025qwen25vl} to extract object names. For inputting the video, we sub-sample the video with 2 fps into frames and ask the model what is held by the hand in the video.

\noindent\textbf{Object mask segmentation.} We use the extracted object names to prompt Grounded-SAM~\cite{ren2024groundedsam} to generate object masks for each frame. A mask confidence threshold of 0.4 and a text threshold of 0.35 are applied. Each object mask is then filtered based on its distance to the hand skeleton and the edge of the hand mask. If multiple objects are detected in a single frame, we select the mask closest to the hand mask. To refine the prompts, we exclude regions corresponding to the hand masks and use the remaining object mask areas from the top three frames with the highest confidence scores to generate point prompts for SAM2, enabling the creation of dense and consistent object masks.


\noindent\textbf{Object Editing.} We use the object masks to generate rectangular masks for each frame, expanding them with an additional margin. A Gaussian dilation with a kernel size of $50\!\times\!50$  px is then applied to these rectangular masks. After excluding the regions corresponding to the hand masks, the resulting final masks are used by VACE-14B~\cite{jiang2025vace} to produce the edited videos. The resolution of the generated videos is $1920\!\times\!1104$ px.

\subsection{Dataset statistics}
\label{appx:dataset_stats}

\begin{figure*}[t] 
    \centering
    
    \begin{subfigure}[b]{0.32\textwidth}
        \centering
        \includegraphics[width=\linewidth]{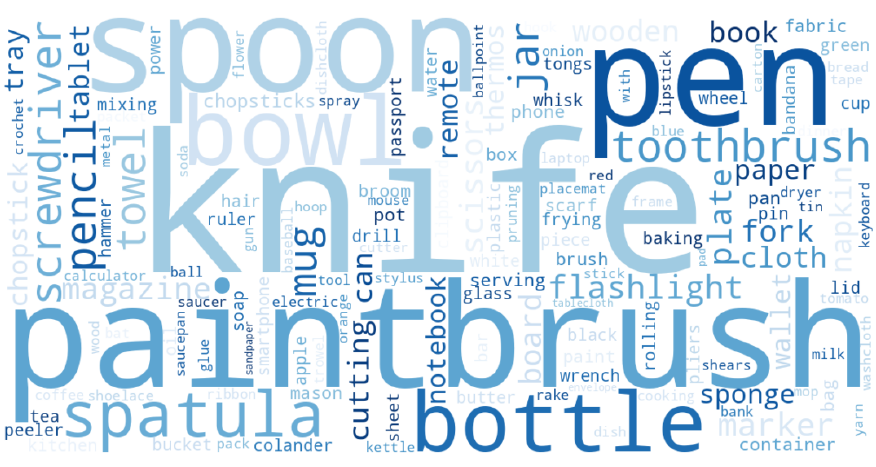}
        \caption{Source Word Cloud} 
        \label{fig:source_wordcloud}
    \end{subfigure}
    \hfill 
    \begin{subfigure}[b]{0.32\textwidth}
        \centering
        \includegraphics[width=\linewidth]{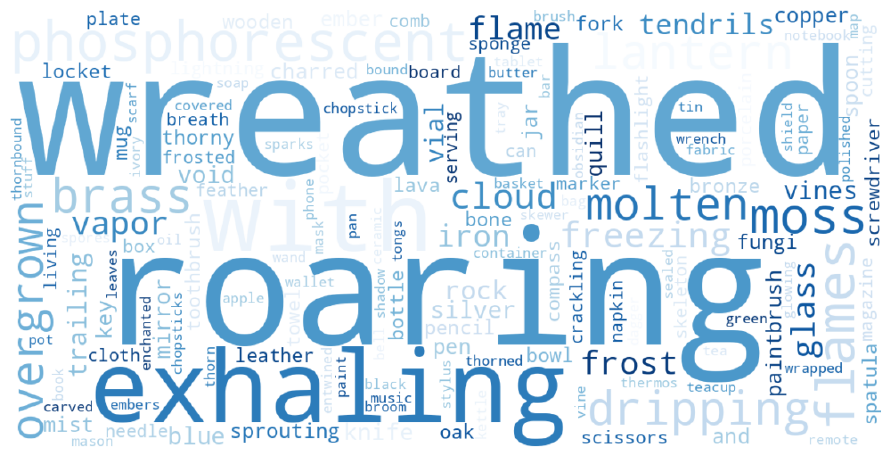}
        \caption{Target Word Cloud} 
        \label{fig:target_wordcloud}
    \end{subfigure}
    \begin{subfigure}[b]{0.32\textwidth}
        \centering
        \includegraphics[width=\linewidth]{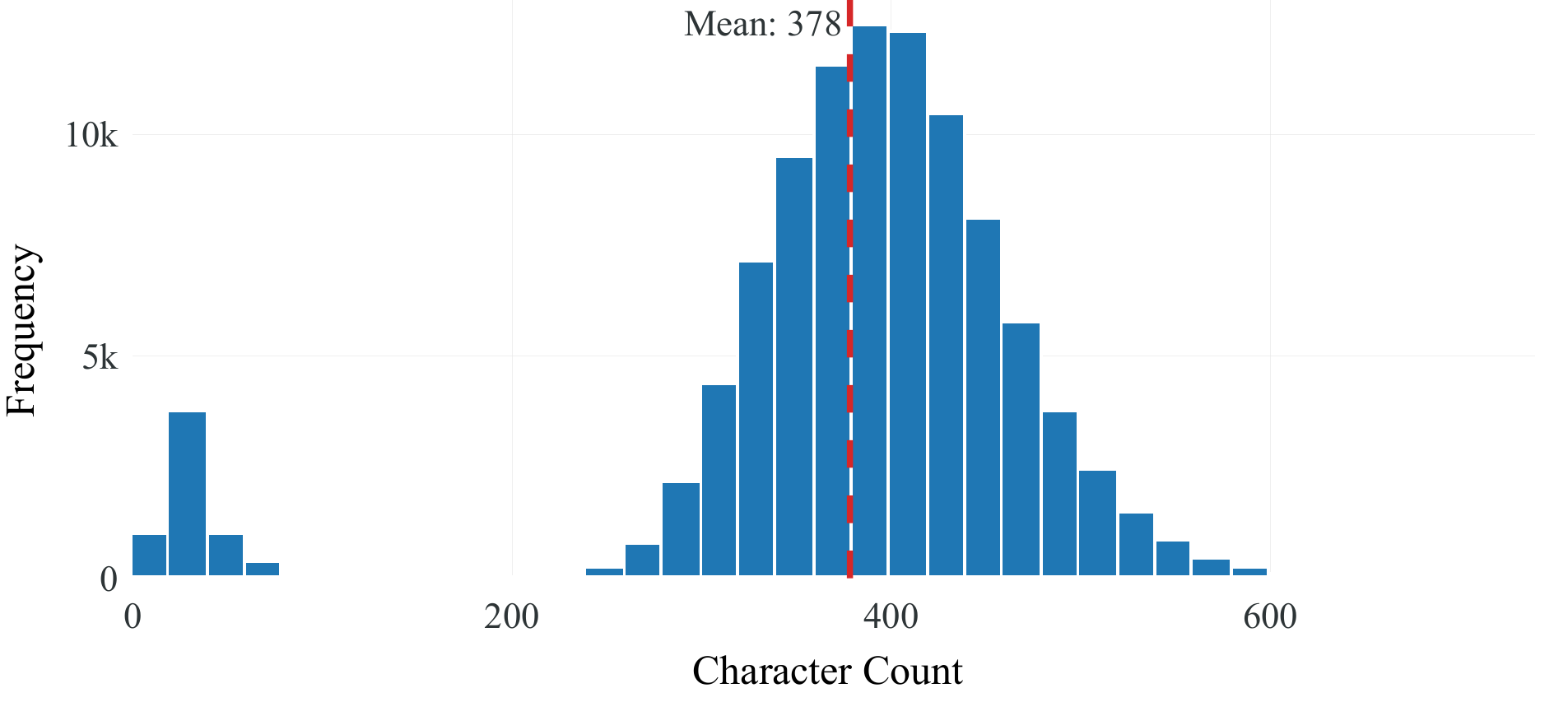}
        \caption{Prompt Length Distribution} 
        \label{fig:prompt_length}
    \end{subfigure}
    
    \caption{Overview of dataset statistics. (a) and (b) illustrate the word clouds for the source and target object descriptions respectively. While source object descriptions focus on everyday objects, target descriptions often contain descriptions of materials and special effects. (c) shows the distribution of prompt lengths. Short prompts in (c) correspond to instructions not requiring context such as \emph{``Remove/Add an \{object\}''}.}
    \label{fig:combined_analysis}
\end{figure*}

\begin{figure}
    \centering
    \includegraphics[width=\linewidth]{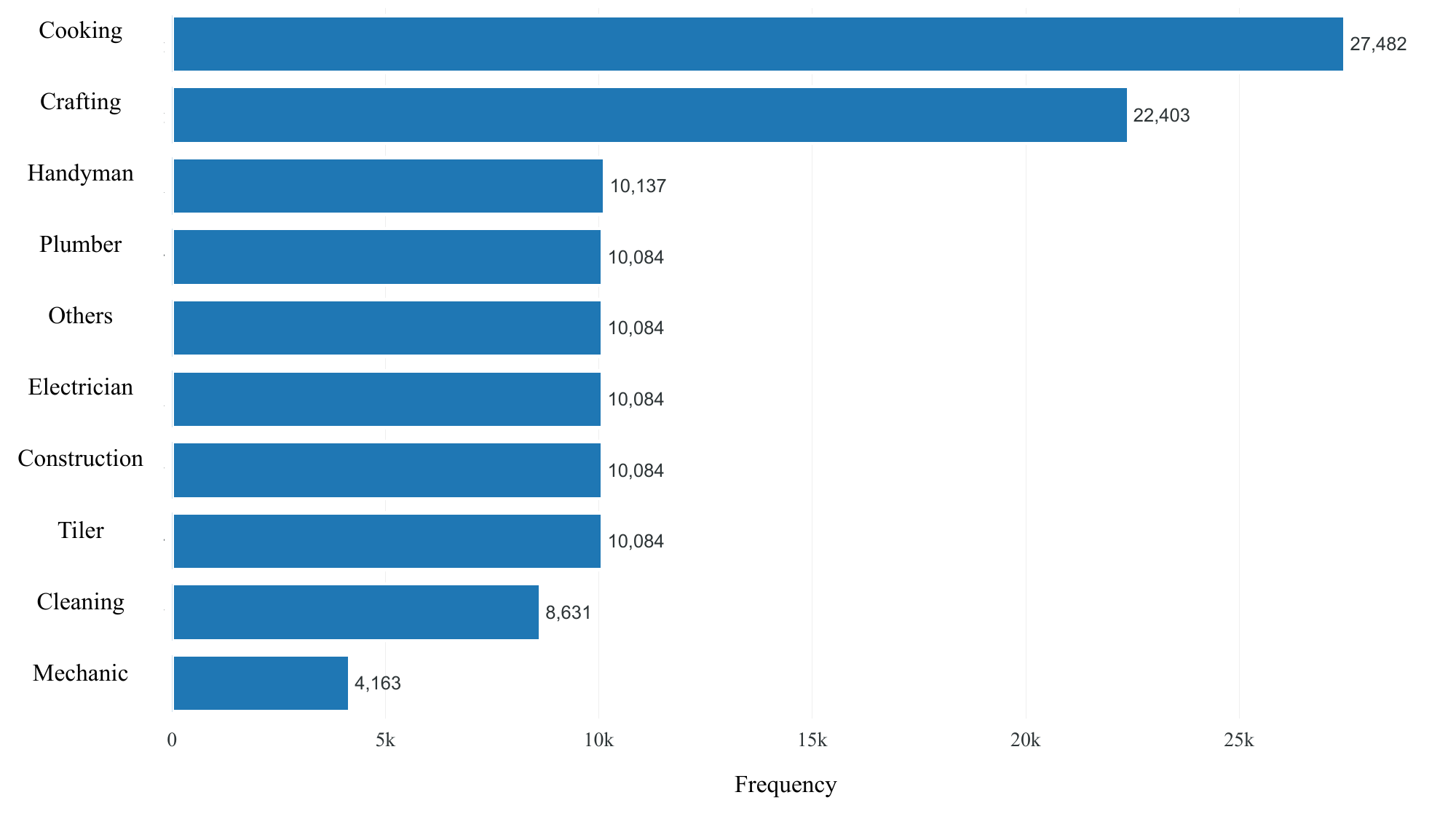}
    \caption{Distribution of most frequent scenarios in \datasetname{} according to the original datasets~\cite{grauman2024egoexo4d,grauman2022ego4d} categorization.}
    \label{fig:top_scenarios}
\end{figure}

\datasetname\ contains approximately $99{,}700$ video editing pairs, each consisting of a source video, a target video, and a corresponding instruction prompt. Our dataset is built upon two egocentric video datasets with $6.4\%$ of the videos originating from EgoExo4D~\cite{grauman2024egoexo4d}, and the remaining $93.6\%$ from Ego4D~\cite{grauman2022ego4d}.
\datasetname{} focuses on object manipulation in egocentric videos. The majority of samples correspond to the \textit{Change Object} task, which prompts the model to replace an existing object in the video with an ordinary object, and contains $54{,}164$ pairs. The second largest group consists of \textit{Change Object with Special Effects} tasks, comprising $39{,}465$ samples, where the source object is replaced with an imaginary object with special effects such as fire or frost. In addition, the dataset includes $3{,}651$ samples for the \textit{Add Object} task and $2{,}379$ for the \textit{Remove Object} tasks.

\Cref{fig:source_wordcloud} presents the word cloud of source objects descriptions, while \Cref{fig:target_wordcloud} shows the word cloud of target objects descriptions. Overall, thanks to the sampling of target object names through GPT-5 Mini~\cite{gpt5} used during the \emph{Object Editing} phase of dataset construction, the dataset contains $13{,}632$ distinct target objects compared to $3{,}199$ unique source objects, enhancing diversity. 
We also report in \Cref{fig:prompt_length} the distribution of the instruction prompt lengths. With a mean of 378 characters, most prompts contain detailed instructions specifying target object, scene details and style, while certain tasks not requiring such context such as \textit{Remove/Add Object} have more concise prompts like \emph{``Remove/Add an \{object\}.''}. 
Finally, \Cref{fig:top_scenarios} shows the ten most frequent scenarios in \datasetname as categorized by the original egocentric datasets, highlighting a balanced distribution across different types of scenes and scenarios.

\section{Additional Method Details}
\label{appx:method_details}

\subsection{Text-to-Video Model}
\label{appx:text_to_video_model}

State-of-the-art methods~\cite{wan,cogvideox,kong2025hunyuanvideo} in text-to-video generation commonly adopt the latent diffusion~\cite{he2023latent} paradigm, DiT-like~\cite{dit} transformer backbones, and the Flow Matching~\cite{liu2022rectifiedflow,lipman2023flowmatching} framework. Our pretrained text-to-video model follows the same paradigm. We model videos in the latent space of a Wan 2.1 autoencoder~\cite{wan} which performs an $8\!\times\!8\!\times\!4$ compression along the height, width, and time dimensions respectively. Input and output linear projections perform $2\!\times\!2$ spatial patchification to further increase compression. The main backbone consists of a 10.7B transformer backbone composed of 32 identical transformer blocks with 4096 hidden dimensions and 32 heads. Each block employs self attention, followed by cross attention for text conditioning, and a final MLP, and uses modulation~\cite{chen2024pixartsigma} for timestep conditioning. QK normalization~\cite{sd3} and Flash Attention~\cite{flash_attention} are used for every attention operation to improve stability and speed. Text tokens are extracted using a combination of pretrained T5~\cite{raffel2022exploring} and CLIP~\cite{clip} text encoders.
The model performs inference in 40 steps using an Euler solver and is trained at a resolution of $512$ px.

\subsection{Training Details}
\label{appx:training_details}

\noindent\textbf{Training configuration.}

We finetune the pretrained text to video model for 30k iterations with a total batch size of 96 videos using 48 H100 GPUs. We use AdamW~\cite{adamw} optimizer with lr 1e-5, a learning rate scheduler with linear warmup of 1000 steps, weight decay of 0.1, beta values of 0.99 and exponential moving average of 0.9999.

\noindent\textbf{Editing data corpora.} 
We collect a corpus of editing data to supplement \datasetname{} for the non-egocentric case.
We consider the following publicly available editing datasets:
GPT-Image-Edit-1.5M~\cite{wang2025gptimageedit},
ShareGPT-4o-Image,
Complex-Edit,
HQEdit,
OmniEdit~\cite{wei2025omniedit}, and
UltraEdit.
In addition, we employ internal data generation pipelines for the creation of additional editing training data. The pipelines employ Qwen-Image-Edit~\cite{wu2025qwenimage} to generate 2M image editing examples, and a combination of models including Wan VACE~\cite{jiang2025vace}, Wan-Animate and MiniMax-Remover to generate an additional corpus of 210k video editing data containing edit categories including object addition and removal, object substitution, human edits. Generated editing pairs are assessed through a combination of automated and manual filtering. In addition, the pipeline synthesizes 1.1M video editing data pair by considering natural videos and pairing them with depth estimation, pose estimation, edge estimation, and optical flow estimation models. The training data is composed of multiple datasets. During training, we apply importance sampling with weights of 28\% for \datasetname{}, 52\% for other video editing datasets, and 20\% for image editing datasets. Within each category, individual datasets are weighted proportionally to their sizes.

\subsection{Distillation Procedure Details}
\label{appx:distillation_procedure}

The distillation procedure is responsible for enabling real-time and low-latency execution of \method{} starting from the finetuned video editing method. The original model performs inference of 5s, $512\!\times\!384$px, 16 fps videos using 40 steps with classifier-free guidance, for a total of 80 model evaluations (NFEs). As every model evaluation takes 86ms on a single H100 GPU, the resulting throughput is of only 11.9fps, which additionally decreases to 9.68fps when considering autoencoder source video encoding and target video decoding times, far from the 16fps required for real-time performance (see \Cref{tab:latency_throughput_performance}).

The first phase of distillation thus performs step and guidance distillation using DMD to yield a model producing a 5s $512\!\times\!384$px 16fps video in 4 NFEs, for a $20\times$ increase in throughput. While the model now possesses a throughput of 43.5fps after application of the autoencoder, the full video needs to be sampled before the first frame can be displayed, and the autoencoder needs to be run on the full source and target videos, creating a first frame latency of 6.93s, inhibiting interactive usage. We perform DMD distillation for 4500 iterations with a total batch size of 64 using 32 H100 GPUs. We use the AdamW~\cite{adamw} optimizer with lr 1e-6 for the generator and 4e-7 for the fake score model, 5 steps per generator update, weight decay of 0.1, beta values of 0.99 and exponential moving average of 0.99.

To enable interactive usage, Self Forcing~\cite{huang2025selfforcing} performs distillation of the bidirectional DMD model into an autoregressive model capable of producing videos in a chunk-by-chunk fashion. In this way, the first frame can be displayed to the user right after the first chunk is sampled and decoded by the autoencoder, reducing first frame latency to 855ms. We use a configuration of Self Forcing where latent frames are denoised in chunks spanning three consecutive latent frames~\cite{huang2025selfforcing}.
We conduct Self Forcing distillation starting directly from the DMD distilled checkpoint for 4500 iterations with a total batch size of 64 using 64 H100 GPUs. We use an AdamW~\cite{adamw} optimizer with lr 1e-6 for the generator and 4e-7 for the fake score model, 10 steps per generator update, weight decay of 0.1, beta values of 0.99 and exponential moving average of 0.99. We find that the model can quickly adapt to autoregressive modeling when initialized from the DMD checkpoint, so we skip the ODE initialization phase~\cite{huang2025selfforcing}. 
We use 7 chunks for generation and each chunk contains 3 latent frames with 21 latent frames in total. Following~\cite{huang2025selfforcing}, to imitate the generation for longer videos, during training, we mask the first chunk of the latent frames when generating the last chunk and we use a window size of 5 chunks as condition during inference.

\section{Additional Benchmark Details}
\label{appx:benchmark_details}

\subsection{Evaluation Tasks}
\paragraph{} Starting from 100 unique egocentric source videos, we construct a benchmark spanning 15 editing tasks. We ensure diversity by clustering source objects and scenario names using sentence embeddings and then uniformly sample source videos across clusters. Conditioned on each source video, its caption, and the source object, we use a GPT-5 Mini~\cite{gpt5} to produce task-specific instruction prompts. Below we include details on constructing the source video and instruction prompt for each editing task.

\begin{itemize}

\item \textbf{Remove Object:} We aim to evaluate the model's ability to remove certain objects from the video while keeping the other parts of the video consistent with the source video. We select 50 source videos from the \datasetname{} and their instruction prompts. We include them in the benchmark and remove them from the training set. 

\item \textbf{Add Object:} We aim to evaluate the model ability to insert a specified object into the scene while keeping the rest of the video consistent with the source. We select 50 source videos from the \datasetname{} together with their instruction prompts. We use source videos from the \emph{Add Object} task, where target object is absent and synthetically removed following the \datasetname{} pipeline. We include these in the benchmark and exclude them from the training set.

\item \textbf{Change Object:} We evaluate the model’s ability to alter a specified object, either by modifying an attribute or replacing it with a new specified object while keeping the rest of the video consistent with the source. For each sampled source video, we create four instruction prompts: two that perform a pure replacement (\emph{object: $a\!\rightarrow\!b$}) and two that pair the replacement with an added effect on the new object (e.g., fire or glow). In total, we obtain 400 videos for evaluating the \textit{Change Object} task. Instruction prompts are designed following \datasetname{} pipeline.

\item \textbf{Change Background:} We evaluate the model’s ability to replace or edit the background while preserving foreground identity and motion. To construct instructions, we provide GPT-5 Mini~\cite{gpt5} with a few source frames and the video caption and request an editing instruction prompt with a semantically compatible target background. 

\item \textbf{Change Camera Pose:} We assess recomposition via a specified camera trajectory (pan/tilt/dolly/zoom) without altering scene events. Instruction prompts are generated by GPT-5~\cite{gpt5} Mini by prompting it with the video caption which contains a description of the camera pose in the source video.

\item \textbf{Add Effect:} We evaluate adding post-processing effects while preserving scene content. These effects usually operate as global filters (e.g., motion blur, VHS, glow, film grain) that are independent of the particular video. Accordingly, we construct instructions by prompting GPT-5 Mini—primed with a few in-context examples from EditVerseBench~\cite{ju2025editverse} and to ask it to propose a diverse pool of effect editing instructions, from which we randomly sample one per clip.

\item \textbf{Stylization:} We mirror the procedure of \textit{Add Effect} by providing GPT-5 Mini with EditVerseBench~\cite{ju2025editverse} examples and ask it to propose diverse set of style editing instructions. We sample one randomly per source video.

\item \textbf{Reasoning:} We evaluate edits that rely on spatial and temporal reasoning. Given the source video and its caption, we prompt GPT-5 Mini~\cite{gpt5} to propose an editing instruction tied to an explicit anchor (event or timestamp) or disambiguating relations (e.g., left-of/behind/before/after). The instruction deliberately avoids explicitly naming a unique target and instead focuses on giving an instruction where the correct object must be inferred from context.

\item \textbf{Depth-to-Video:}~We convert the source video into a depth map using Depth Anything~\cite{depthanything}, and construct the instruction prompt as (\emph{``Turn the depth map into a video with the following description: {caption}.''}) where the {caption} is the source video caption describing the appearance of the scene. 
\item \textbf{Sketch-to-Video:}~We convert the source video into Canny edge maps using OpenCV, and construct the instruction prompt as (\emph{``Turn the canny edge map into a video with the following description: {caption}.''}), where {caption} is the source video caption describing the scene’s appearance and layout.

\item \textbf{Pose-to-Video:}~We convert the source video into 2D human poses using DWpose~\cite{yang2023dwpose}, and construct the instruction prompt as (\emph{``Turn the DWpose pose map into a video with the following description: $\langle$caption$\rangle$.''}), where {caption} is the source video caption specifying the subject’s identity, attire, and overall scene appearance.

\item \textbf{Video-to-Depth:}~We prompt the model to convert a video into a temporally consistent depth map. For this task, we use fixed instruction prompt (\emph{``Turn the video into a depth map.''}).

\item \textbf{Video-to-Sketch:}~We use a fixed instruction prompt of (\emph{``Turn the video into a Canny edge map.''}).

\item \textbf{Video-to-Pose:}~We use a fixed instruction prompt (\emph{``Turn the video into a DWpose pose map.''}).

\item \textbf{Combined (Multi-Task):}~We compose multiple editing prompts from the same source video (e.g., Pose-to-Video + Change Background + Stylization) by sampling a subset of instruction prompts and prompting a GPT-5 Mini~\cite{gpt5} to compose a single instruction prompt that combines all of the tasks together.
\end{itemize}

\subsection{EgoEditBench-Human Alignment}
\begin{table*}[t]
\centering

\setlength{\tabcolsep}{6pt}
\renewcommand{\arraystretch}{1.15}

\resizebox{\textwidth}{!}{%
\rowcolors{2}{gray!8}{white}%
\begin{tabular}{l ccc cc cc cc}
\toprule
& \multicolumn{3}{c}{\textbf{VLM Mean Score}} & \multicolumn{2}{c}{\textbf{Preference (VLM)}} & \multicolumn{2}{c}{\textbf{Preference (User Study)}} & \multicolumn{2}{c}{\textbf{Agreement (\%)}} \\
\cmidrule(lr){2-4}\cmidrule(lr){5-6}\cmidrule(lr){7-8}\cmidrule(lr){9-10}
\textbf{Task} & \textbf{\method} & \textbf{LucyEdit} & \textbf{InsV2V} & \textbf{vs LucyEdit} & \textbf{vs InsV2V} & \textbf{vs LucyEdit} & \textbf{vs InsV2V} & \textbf{LucyEdit} & \textbf{InsV2V} \\
\midrule
Add Object              & 7.83 & 4.12 & 5.47 & 29 (100\%) & 29 (97\%) & 29 (100\%) & 29 (97\%) & 100.0 & 93.3 \\
Change Camera Pose      & 7.11 & 6.93 & 7.30 & 20 (67\%)  & 14 (47\%) & 27 (90\%)  & 28 (93\%) & 70.0  & 40.0 \\
Change Object           & 7.61 & 6.49 & 4.09 & 23 (77\%)  & 26 (87\%) & 29 (97\%)  & 30 (100\%)& 80.0  & 86.7 \\
Change Background       & 7.28 & 5.10 & 4.46 & 29 (97\%)  & 27 (90\%) & 28 (93\%)  & 29 (97\%) & 90.0  & 86.7 \\
Combined (Multi-Task)   & 7.96 & 6.36 & 4.66 & 30 (100\%) & 29 (97\%) & 29 (97\%)  & 30 (100\%)& 96.7  & 96.7 \\
Depth-to-Video          & 8.53 & 7.44 & 5.72 & 30 (100\%) & 30 (100\%)& 30 (100\%) & 30 (100\%)& 100.0 & 100.0 \\
Add Effect              & 6.41 & 5.29 & 5.96 & 24 (80\%)  & 19 (63\%) & 24 (80\%)  & 27 (90\%) & 80.0  & 73.3 \\
Video-to-Pose           & 7.90 & 4.63 & 4.39 & 30 (100\%) & 30 (100\%)& 28 (93\%)  & 28 (93\%) & 93.3  & 93.3 \\
Pose-to-Video           & 8.58 & 7.18 & 4.77 & 28 (93\%)  & 30 (100\%)& 28 (93\%)  & 29 (97\%) & 86.7  & 96.7 \\
Reasoning               & 6.69 & 5.30 & 4.85 & 24 (80\%)  & 23 (77\%) & 25 (83\%)  & 29 (97\%) & 63.3  & 73.3 \\
Remove Object           & 6.72 & 5.04 & 5.71 & 25 (83\%)  & 23 (77\%) & 26 (87\%)  & 29 (97\%) & 76.7  & 73.3 \\
Sketch-to-Video         & 8.81 & 6.31 & 5.54 & 30 (100\%) & 30 (100\%)& 27 (90\%)  & 30 (100\%)& 90.0  & 100.0 \\
Stylization             & 7.88 & 4.68 & 6.70 & 29 (97\%)  & 25 (83\%) & 29 (97\%)  & 29 (97\%) & 93.3  & 80.0 \\
Video-to-Depth          & 8.80 & 4.00 & 4.80 & 30 (100\%) & 30 (100\%)& 28 (93\%)  & 28 (93\%) & 93.3  & 93.3 \\
Video-to-Sketch         & 9.00 & 3.78 & 4.34 & 30 (100\%) & 30 (100\%)& 24 (80\%)  & 26 (87\%) & 80.0  & 86.7 \\
\midrule
Overall                 &  7.76    &  5.44    &  5.24    & 411 (91\%) & 395 (88\%)&  411 (91\%)& 431 (96\%)& 86.2  & 84.9 \\  
\bottomrule

\end{tabular}%
}
\caption{Study on \benchmarkname{} comparing VLM and human preference alignment. We conduct VLM evaluation following \benchmarkname{} protocol and for each sample, assign VLM preference to the method with highest VLM score. Simultaneously we ask a human evaluator to express preference for a sample from one of two compared methods. We find VLM and user preference to be in agreement in 86.2\% and 84.9\% of cases on average, respectively when evaluated against LucyEdit and InsV2V.}
\label{tab:method_lucy_insv2v_agreement}
\end{table*}
To evaluate the reliability of the VLM score employed in \benchmarkname{}, we conduct a study on \benchmarkname{} comparing VLM and human preference alignment. Given 30 randomly sampled benchmark element per category, and a baseline method, we conduct VLM evaluation of \method{} and the baseline following \benchmarkname{} protocol and, for each sample, assign VLM preference to the method with highest VLM score. Simultaneously we ask a human evaluator to express preference for each sample for our method or the baseline. Results are shown in \Cref{tab:method_lucy_insv2v_agreement}. VLM and user preferences are in high agreement, with 86.2\% and 84.9\% of cases on average, respectively when evaluated against LucyEdit and InsV2V. We thus rely on VLM score as the main benchmark metric.

\section{Evaluation Details}
\label{appx:evaluation_details}

\subsection{Baseline Details}
\label{appx:baseline_details}

\noindent To ensure a fair comparison, we use each baseline’s default inference hyperparameters including the number of frames, frames-per-second (FPS), spatial resolution, and inference settings such as guidance scale and sampling steps. Below, we include the specific hyperparameters we used for evaluating each baseline:

\begin{itemize}
\item \textbf{TokenFlow.} We rely on Stable Diffusion 1.5 as the backbone. We use $16$ frames at a 
\item \textbf{STDF.} We use $24$ frames at $10$ fps and 40 inference steps and 10 optimization steps. We use a guidance scale of $10$ and inference at the resolution of $576\times320$. 
 
\item \textbf{SENORITA (Senorita-2M).} We use $33$ frames at $8$ fps and $768\times448$ resolution, with guidance scale of $4$ and $30$ inference steps. We use CogVideoX as the backbone. To obtain the edited first frame, we use the first frame generated by our model instead of relying on pretrained ControlNets, following~\cite{ju2025editverse}. We crop and resize the first frame generated by our method to the default resolution of SENORITA.

\item \textbf{AnyV2V.} We use $16$ frames at $8$ fps and $512\times512$, guidance of $9$, with $100$ inversion steps plus $50$ edit steps. Similarly, we use the first frame generated by our model instead of relying on pretrained ControlNets, following~\cite{ju2025editverse}. We crop and resize the first frame generated by our method to the default resolution of AnyV2V.

\item \textbf{InsV2V.} We use $32$ frames at $15$ fps and $384\times384$, with the default guidance for the video branch ($1.2$) and text branch ($7.5$), over $20$ inference steps.

\item \textbf{Lucy-Edit.} We use $81$ frames at $15$ fps with $832\times480$ resolution, guidance scale of $5$, and $50$ inference steps.

\item \textbf{EditVerse.} Since EditVerse is a closed source model, we only compare on EditVerseBench, where we rely on its published samples. We omit the results on \benchmarkname{} since we do not have access to the model to generate the required results.

\item \textbf{StreamDiffusion.} We use the image-to-image editing pipeline for the video editing task. We process $81$ frames and $16$ fps at the resolution $832\times480$. 

\item \textbf{StreamDiffusionV2.} We use the streaming setup at $832\times480$ with an $81$ frames sequence and $16$ fps, using a very light $2$–$4$ denoising steps.

\end{itemize}
\subsection{Metrics Details}
\noindent We closely follow \cite{ju2025editverse} when computing metrics for our benchmark. We evaluate edit quality with a VLM-based judge, overall video quality with PickScore, text alignment at the frame and video levels, and temporal consistency with CLIP and DINO. Because frame- and video-level text alignment are highly correlated in our setting, we report only the video-level score for brevity. We also find the CLIP- and DINO-based consistency metrics to be strongly correlated, so we rely on the CLIP consistency metric for temporal consistency. Among all metrics, we found the VLM score to align well with human judgment (see \Cref{tab:method_lucy_insv2v_agreement}), so we use it as our primary quality signal. For EgoEditBench, we use exactly the same evaluation prompts as \cite{ju2025editverse} when computing the VLM editing-quality score.

\section{Additional Results}

\begin{table*}[t]
\centering

\resizebox{\textwidth}{!}{%
\rowcolors{2}{gray!8}{white}%
\begin{tabular}{lcccccccccccccccc}
\toprule
\textbf{Method} & \rotatebox{45}{\textbf{Add}} & \rotatebox{45}{\textbf{Cam. Mov.}} & \rotatebox{45}{\textbf{Change}} & \rotatebox{45}{\textbf{Chg. BG}} & \rotatebox{45}{\textbf{Combined}} & \rotatebox{45}{\textbf{Depth2Vid}} & \rotatebox{45}{\textbf{Effect}} & \rotatebox{45}{\textbf{Vid2Pose.}} & \rotatebox{45}{\textbf{Pose2Vid}} & \rotatebox{45}{\textbf{Reasoning}} & \rotatebox{45}{\textbf{Remove}} & \rotatebox{45}{\textbf{Sketch2Vid}} & \rotatebox{45}{\textbf{Stylization}} & \rotatebox{45}{\textbf{Vid2Depth}} & \rotatebox{45}{\textbf{Vid2Sketch}} & \rotatebox{45}{\textbf{Overall}} \\
\midrule
TokenFlow           & 5.56 & 5.67 & 5.37 & 5.02 & 5.39 & 4.05 & 5.80 & 5.62 & 0.52 & 5.24 & 5.49 & 3.73 & 6.01 & 5.88 & 5.54 & 4.99 \\
STDF                & 4.59 & 4.40 & 4.83 & 4.82 & 5.27 & 4.71 & 4.67 & 5.33 & 2.58 & 4.13 & 4.55 & 4.00 & 5.33 & 5.17 & 4.45 & 4.59 \\
Se\~norita-2M       & 7.85 & 7.34 & 7.35 & 6.64 & 7.10 & \underline{8.48} & 7.24 & 6.68 & \underline{7.85} & 7.61 & 6.93 & \underline{8.49} & 6.75 & 8.41 & 8.05 & 7.52 \\
AnyV2V              & 7.72 & 7.55 & 7.22 & 6.58 & 6.42 & 7.14 & \textbf{7.32} & 6.66 & 2.62 & 7.52 & \textbf{7.21} & 5.29 & 6.58 & 7.35 & 7.56 & 6.72 \\
InsV2V              & 5.67 & 7.26 & 3.99 & 4.42 & 4.85 & 5.60 & 6.30 & 4.39 & 4.69 & 4.41 & 5.52 & 5.55 & 6.73 & 4.79 & 4.40 & 5.22 \\
Lucy Edit           & 4.31 & 6.92 & 6.25 & 4.67 & 6.15 & 7.31 & 5.16 & 4.49 & 7.13 & 5.68 & 4.81 & 6.52 & 4.65 & 3.88 & 3.72 & 5.46 \\
StreamDiffusion     & 3.95 & 3.06 & 3.44 & 5.29 & 5.60 & 5.53 & 3.62 & 5.43 & 2.66 & 3.26 & 2.91 & 5.20 & 6.32 & 4.45 & 4.02 & 4.31 \\
StreamDiffusionV2   & 1.63 & 1.76 & 2.42 & 2.21 & 3.62 & 4.66 & 2.42 & 3.52 & 2.45 & 1.10 & 1.02 & 3.33 & 3.19 & 3.53 & 1.37 & 2.55 \\
\midrule
EgoEdit             & \textbf{7.89} & 6.79 & 7.84 & \textbf{7.45} & \textbf{7.74} & \textbf{8.57} & 6.32 & 7.87 & \textbf{8.57} & 6.79 & 6.82 & \textbf{8.70} & \textbf{7.46} & \textbf{8.70} & \textbf{8.93} & \textbf{7.76} \\
EgoEdit-DMD         & 7.91 & 6.76 & \underline{8.04} & 5.36 & 7.29 & 8.01 & 5.92 & \underline{8.07} & 5.60 & 7.16 & \underline{7.46} & 7.76 & 6.45 & \underline{8.95} & \underline{8.96} & 7.42 \\
EgoEdit-RT          & \underline{7.79} & \underline{6.67} & \textbf{8.09} & \underline{7.57} & \underline{7.83} & \underline{8.52} & \underline{6.40} & \textbf{8.48} & \underline{8.31} & \underline{7.54} & 4.01 & \underline{7.92} & \underline{7.43} & 8.89 & 8.98 & \underline{7.83} \\
\bottomrule
\end{tabular}
}%
\caption{Breakdown of per-category \benchmarkname{} VLM scores for \method{} and baselines.}
\label{tab:all_baselines_vlm_renamed_ordered}
\end{table*}
\renewcommand{\oursrowcolor}{azure!6}
\renewcommand{\pickscorefull}{Pick Score\xspace}
\renewcommand{\pickscoreshort}{PS\xspace}
\renewcommand{\vlmevalfull}{VLM evaluation\xspace}
\renewcommand{\vlmevalshort}{VLM\xspace}

\begin{table}[t]
\centering
\small
\rowcolors{2}{gray!8}{white}%
\begin{tabular}{lcccc}
\toprule
\textbf{Method}
& \textbf{\vlmevalshort} $\uparrow$ & \textbf{\pickscoreshort} $\uparrow$ & \textbf{TA} $\uparrow$ & \textbf{TC} $\uparrow$ \\
\midrule
\textbf{EgoEdit} & 7.80 & \cellfirst{19.09} & \cellfirst{16.91} & 96.74 \\
\textbf{EgoEdit-DMD} & 7.42 & 18.95 & 16.52 & \cellfirst{96.87} \\
\textbf{EgoEdit-RT} & \cellfirst{7.83} & 19.04 & 16.49 & 96.49 \\
\bottomrule
\end{tabular}
\caption{Quantitative comparison of \method variants on EgoEditBench: 
``VLM'' is VLM evaluation score, ``PS'' is Pick Score, ``TA'' is Text Alignment, 
``TC'' is Temporal Consistency. EgoEdit-RT is the real-time version, and EgoEdit-DMD 
is the bidirectional DMD variant.}
\label{tab:ablation_distillation_supp}
\end{table}
\begin{table*}[t]
\centering

\resizebox{\textwidth}{!}{%
\rowcolors{2}{gray!8}{white}%
\begin{tabular}{lcccccccccccc}
\toprule
\textbf{\% of EgoEditData} & \rotatebox{45}{\textbf{Add}} & \rotatebox{45}{\textbf{Cam. Mov.}} & \rotatebox{45}{\textbf{Change}} & \rotatebox{45}{\textbf{Chg. BG}} & \rotatebox{45}{\textbf{Combined}} & \rotatebox{45}{\textbf{Depth2Vid}} & \rotatebox{45}{\textbf{Pose2Vid}} & \rotatebox{45}{\textbf{Reasoning}} & \rotatebox{45}{\textbf{Remove}} & \rotatebox{45}{\textbf{Sketch2Vid}} & \rotatebox{45}{\textbf{Stylization}} & \rotatebox{45}{\textbf{Overall}} \\
\midrule
0\%                   & 7.80 & 6.27 & 6.38 & \textbf{8.33} & 5.54 & 7.63 & 8.17 & 6.43 & 4.30 & 7.40 & 7.87 & 6.89 \\
25\%                  & 7.23 & 6.43 & 6.82 & 8.23 & \textbf{7.42} & 6.00 & 8.63 & 6.73 & 5.00 & 6.63 & 8.20 & 7.00 \\
75\%                  & 7.73 & 7.37 & 6.88 & 7.80 & 5.29 & 8.07 & \textbf{8.63} & 7.53 & 5.27 & \textbf{7.53} & 8.60 & 7.30 \\
100\%                 & \textbf{7.92} & \textbf{7.60} & \textbf{7.46} & 8.27 & 7.17 & \textbf{8.13} & 8.57 & \textbf{8.37} & \textbf{7.00} & 7.17 & \textbf{8.77} & \textbf{7.79} \\
\bottomrule
\end{tabular}
}%
\caption{\textbf{EditVerseBench~\cite{ju2025editverse}} results for our model trained on a data mixture where the number of samples retained in \datasetname{} is varied. As the amount of retained samples in \datasetname{} increases, non-egocentric editing performance as measured by EditVerseBench~\cite{ju2025editverse} increases, showing usefulness of \datasetname{} beyond the egocentric case.}
\label{tab:ablation_egoeditdata}
\end{table*}
\label{appx:additional_results}

\subsection{Additional In-the-Wild Results}
\begin{figure*}
    \centering
    \includegraphics[width=\linewidth]{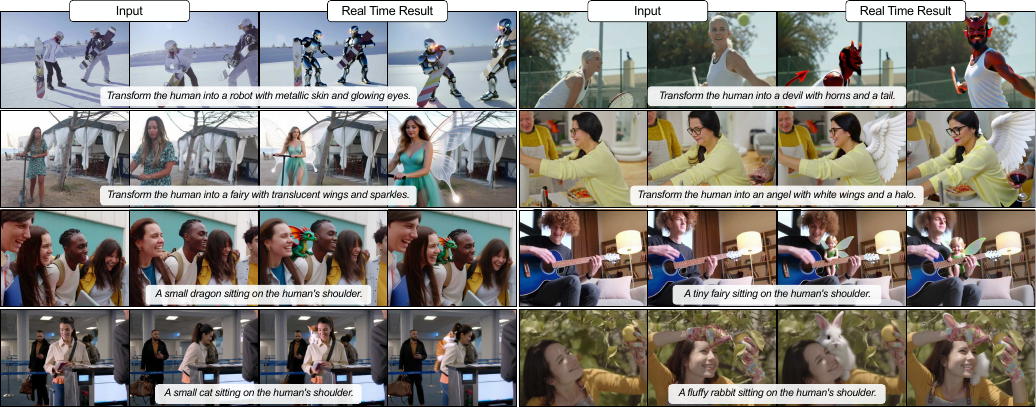}
    \caption{Exocentric video edits produced in real time by \method's streaming variant \methodsf{} on a single H100 GPU.}
    \label{fig:qualitatives_exocentric_supplement}
\end{figure*}

\begin{figure*}[p]
    \centering
    \includegraphics[width=\linewidth]{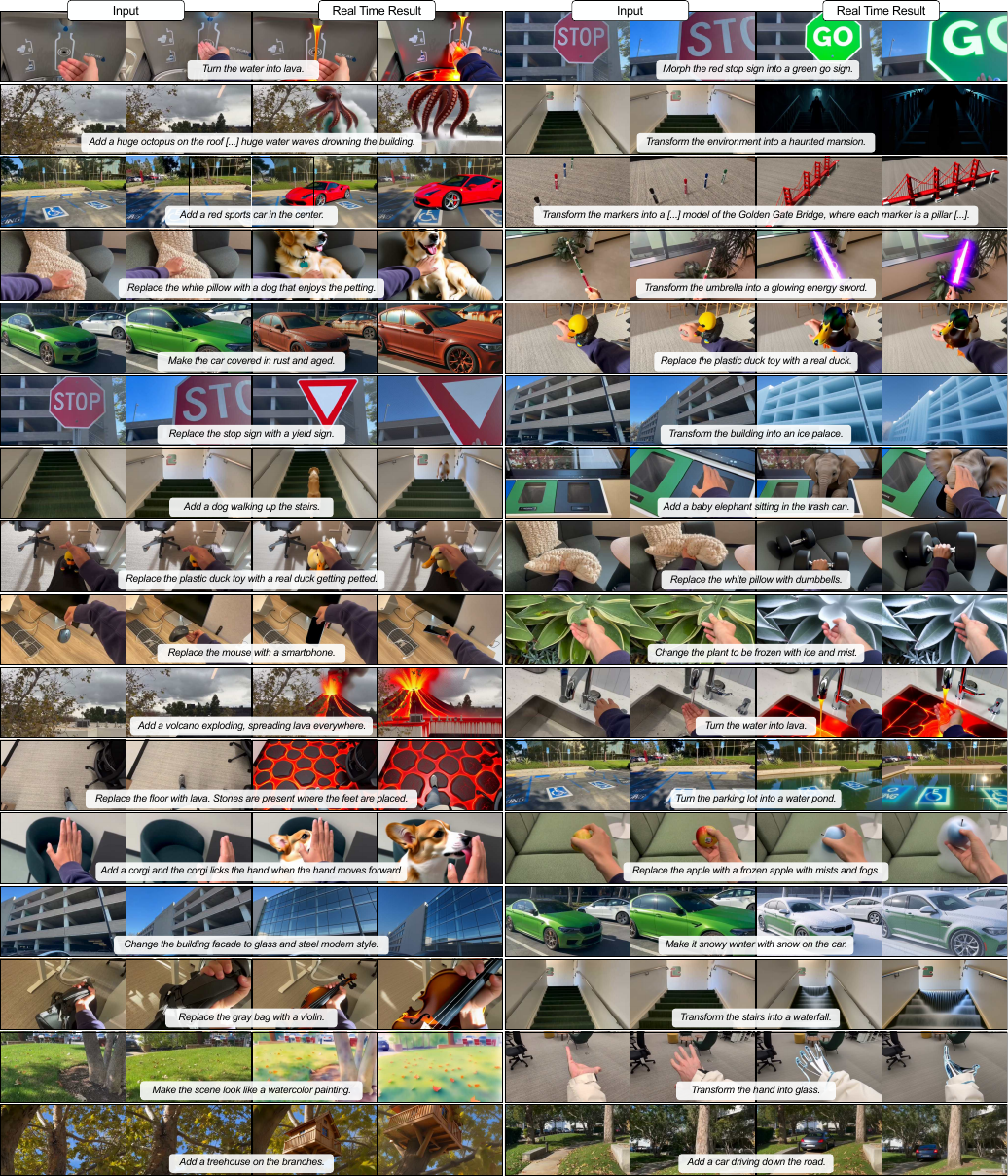}
    \caption{In-the-wild video edits produced in real time by \method's streaming variant \methodsf{} on a single H100 GPU.}
    \label{fig:qualitatives_in_the_wild_supplement}
\end{figure*}

We present in \Cref{fig:qualitatives_in_the_wild_supplement} and \Cref{fig:qualitatives_exocentric_supplement} additional in-the-wild qualitative examples produced by \methodsf{} in real-time on a single H100 GPU for the egocentric and exocentric cases respectively. \methodsf{} can perform complex edits such as transforming markers placed on the floor into pillars of the Golden Gate Bridge, model fluid effects, change a location into a haunted mansion, add animals that interact with the environment and the user, or perform stylization.

\subsection{Additional Comparison to Baselines}
\begin{figure*}
    \centering
    \includegraphics[width=\linewidth]{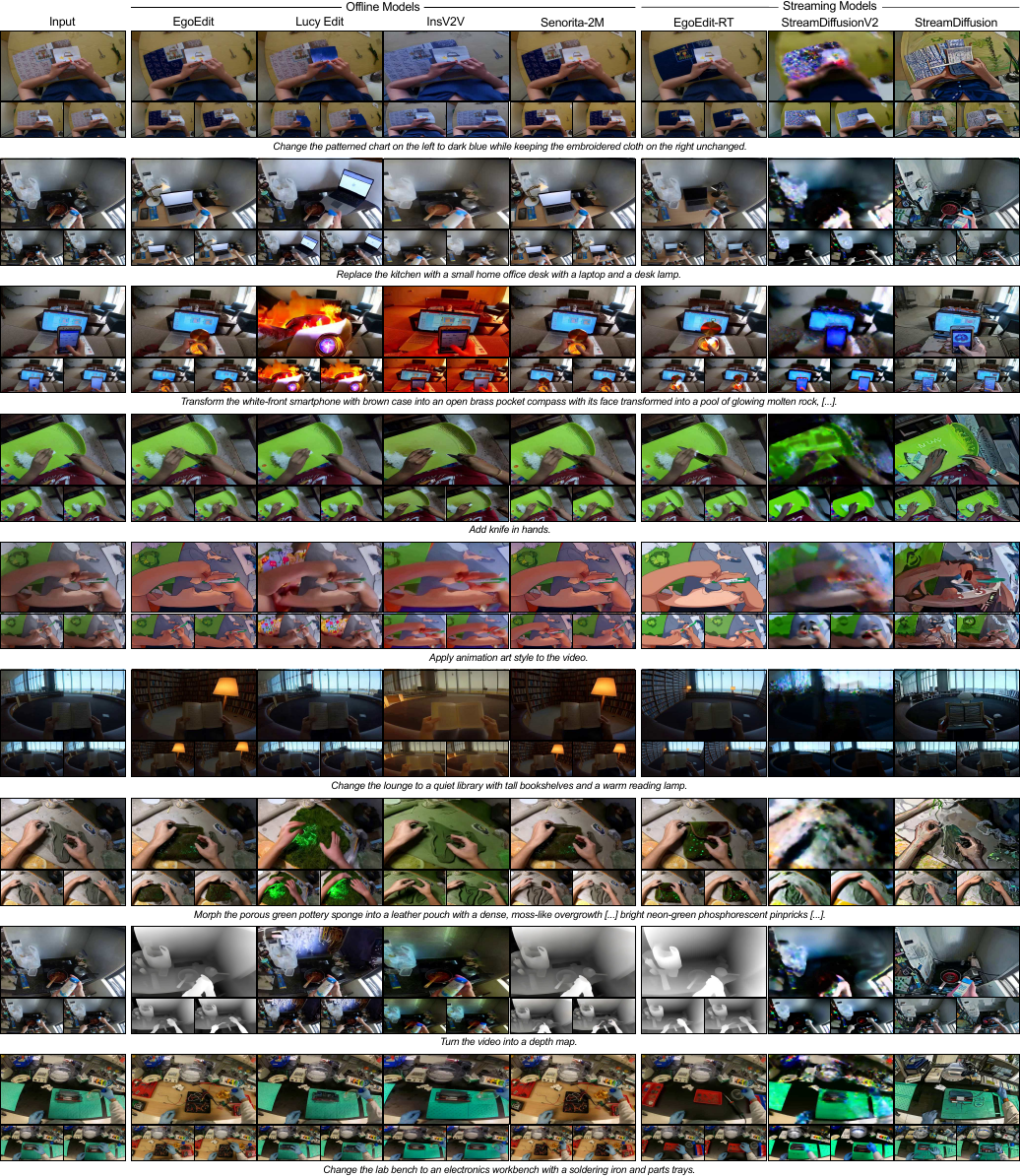}
    \caption{Qualitative comparison of \method{} and its real-time streaming variant \methodsf{} against baselines on \benchmarkname{}. \method and \methodsf consistently perform better than their baselines. Note that Señorita-2M uses the first frame from \method for frame propagation.}
    \label{fig:qualitatives_baselines_supplement}
\end{figure*}

We present additional qualitative results in \Cref{fig:qualitatives_baselines_supplement} comparing \method{} to baseline video editing methods. \method{} and its real-time variant \methodsf{} are capable of performing a range of editing tasks including object attribute changes, object replacement, object insertion, style transfer, background changes, and conversion to depth map. We observe concurrent offline methods~\cite{lucyedit2025} and real-time video editing methods~\cite{feng2025streamdiffusionv2,kodaira2025streamdiffusion} to often fail in the egocentric case, with failure modes often manifesting as inability to produce any change over the source video, or modifying the input beyond what is requested in the editing instruction. In addition, \Cref{tab:all_baselines_vlm_renamed_ordered} reports per-category VLM scores on \benchmarkname{} for our method and baselines.

\subsection{Additional Distillation Ablation Results}
\begin{figure*}
    \centering
    \includegraphics[width=\linewidth]{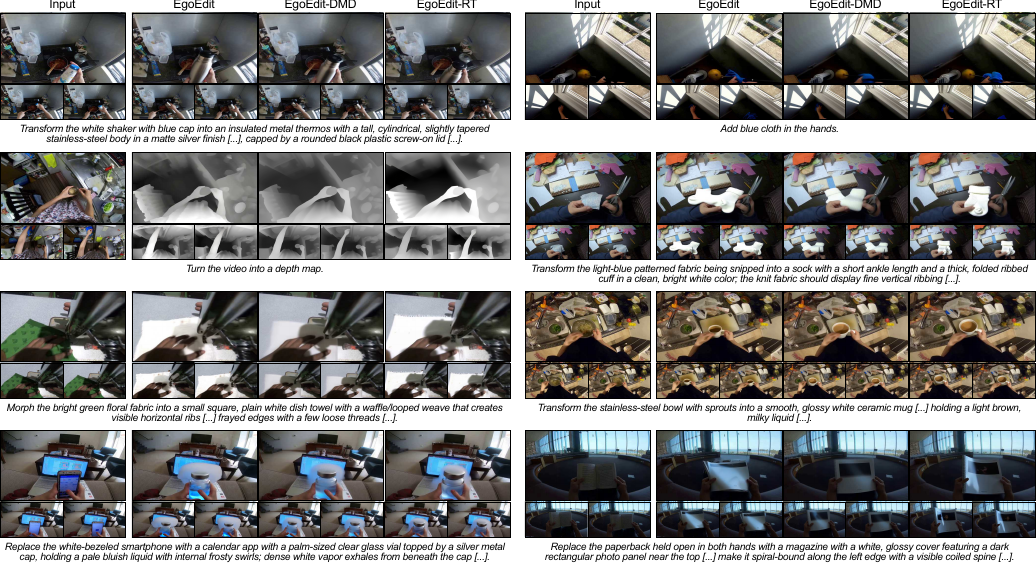}
    \caption{Qualitative comparison of \method{} at different stages of distillation. \method{} indicates the original 80 NFEs model, \methoddmd{} represents the model after the 4-step DMD distillation, \methodsf{} represents the final real-time streaming model obtained after Self Forcing distillation.}
    \label{fig:qualitatives_distillation_supplement}
\end{figure*}

\Cref{fig:qualitatives_distillation_supplement} shows additional qualitative results comparing variants of \method{} at different stages of distillation: the starting 40-step (80 NFEs) video editing checkpoint obtained after finetuning of the pretrained video generation model (\method{}), the 4-step (4 NFEs) variant obtained as a result of the DMD distillation phase (\methoddmd{}), and the real-time streaming variant obtained at the end of Self Forcing training (\methodsf{}). We observe similar qualitative performance among the distilled variants as shown in the figure. We observe \methodsf{} to occasionally present temporal shifts at the boundaries between different chunks of predicted frames. In addition, when qualitatively evaluated on in-the-wild cases, distilled variants (\methoddmd{}, \methodsf{}) exhibited a lower capacity of handling complex out-of-distribution editing instructions. In addition, \Cref{tab:ablation_distillation_supp} reports full evaluation scores on \benchmarkname{} for all distilled variants of \method{}.

\subsection{Additional Dataset Ablation Results}
\begin{figure*}
    \centering
    \includegraphics[width=\linewidth]{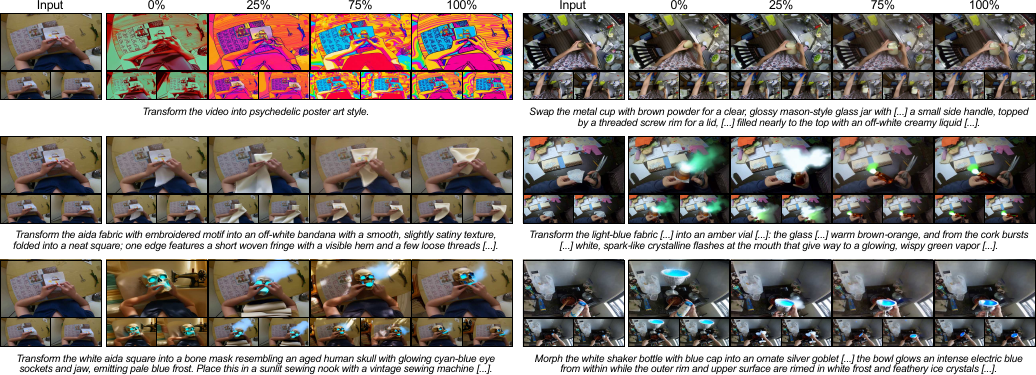}
    \caption{Qualitative comparison of different variants of \method{} trained on a data mixture with reduced portions of \datasetname{}. Percentages indicate the proportion of unique source video samples in \datasetname{} retained for training.}
    \label{fig:qualitatives_dataset_ablation_supplement}
\end{figure*}

In \Cref{fig:qualitatives_dataset_ablation_supplement}, we show qualitative examples produced by different variants of \method{} trained on our training data mixture same as the main experiment (\Cref{tab:egobench}) with reduced portions of \datasetname{}. As the portion of \datasetname{} increases, we notice increase quality of edits and improved alignment to the source video. We additionally evaluate whether increasing the amount of data in \datasetname{} can result in improved performance in non-egocentric benchmarks. \Cref{tab:ablation_egoeditdata} compares variants of \method{} trained with a data mixture same as the main experiment including versions of \datasetname{} of reduced size, on the EditVerseBench benchmark, which mostly consists of exocentric videos. We find that the introduction of \datasetname{} consistently improves the overall benchmark score, with particular regards to the \emph{Reasoning}, \emph{Remove}, \emph{Camera Movement}, and \emph{Change} categories. We speculate these improvements originate from strict quality filtering and descriptive captions, whose benefits extend beyond the egocentric case.

\section{Failed Experiments}
\label{appx:failed_experiments}

\noindent\textbf{Usage of unfiltered data.} We initially experiment with a video editing dataset corpora consisting of lightly filtered video editing pairs. In this setting, we notice weak video editing performance, with the model often reproducing artifacts encountered in low-quality video editing data pairs such as failure of an object in being replaced with a different object, or failure in adding an object. This motivates us to integrate extensive automated and manual curation into \datasetname{} at the expense of dataset size, which resulted in increased editing performance.

\noindent\textbf{Usage of detailed editing instructions.} We initially experiment with simple editing instructions for \datasetname{} that are obtained by filling templates with known source and target object names such as \emph{``Replace $\langle$source object$\rangle$ with $\langle$target object$\rangle$''}, where source and target objects are represented by strings obtained during the \emph{Object names extraction} and \emph{Object Editing} stages of the data curation pipeline.
We observe low editing performance when training on such prompts.
Such captions suffer from several issues: (i) their variety is limited, (ii) object names extracted in the \emph{Object names extraction} stage are often short, (iii) objects in generated target videos produced in the \emph{Object Editing} stage might not faithfully correspond to the requested edited object description. After training on editing instructions using GPT-5 Mini~\cite{gpt5}, we observe increased ability to follow instruction prompts.

\end{document}